\documentclass[final,3p,times]{elsarticle}

\usepackage{amssymb}
\usepackage{amsmath}

\usepackage{multirow}
\usepackage{adjustbox}
\usepackage{tabularx}
\usepackage{booktabs}
\usepackage{graphicx}
\usepackage{array}
\newcolumntype{M}[1]{>{\centering\arraybackslash}m{#1}}
\newcolumntype{J}[1]{>{\raggedright\arraybackslash}m{#1}} 
\usepackage{newtxtext, newtxmath, xcolor, colortbl}
\usepackage{threeparttable}
\usepackage{pifont}
\journal{Applied Soft Computing}
\usepackage{hyperref}
\usepackage{xcolor}

\begin{document}

\begin{frontmatter}



\title{Proxy Reconstruction Pre-training for Ramp Flow Prediction at Highway Interchanges}


\author[seut]{Yongchao Li}
\ead{230249216@seu.edu.cn}

\author[seut,spu]{Jun Chen}
\ead{chenjun@seu.edu.cn}

\author[seum,ippt]{Zhuoxuan Li}
\ead{230229338@seu.edu.cn}

\author[itssky]{Chao Gao}
\ead{gaochao@itssky.com}

\author[seut,cityu]{Yang Li}
\ead{l_yang@seu.edu.cn}

\author[seut,jpcic]{Chu Zhang\corref{cor1}}
\ead{zhangchu0720@seu.edu.cn}
\cortext[cor1]{Corresponding author}

\author[nwpu,nklacd]{Changyin Dong\corref{cor1}}
\ead{dongcy@nwpu.edu.cn}

\affiliation[seut]{
  organization={School of Transportation, Southeast University},
  city={Nanjing},
  postcode={211189},
  country={China}
}

\affiliation[jpcic]{
  organization={Jiangsu Province Collaborative Innovation Center of Modern Urban Traffic Technologies},
  city={Nanjing},
  postcode={211189},
  country={China}
}

\affiliation[nwpu]{
  organization={School of Aeronautics, Northwestern Polytechnical University},
  city={Xi’an},
  postcode={710071},
  country={China}
}

\affiliation[nklacd]{
  organization={National Key Laboratory of Aircraft Configuration Design},
  city={Xi’an},
  postcode={710071},
  country={China}
}

\affiliation[seum]{
  organization={School of Mathematics, Southeast University},
  city={Nanjing},
  postcode={211189},
  country={China}
}

\affiliation[itssky]{
  organization={ITSSKY Technology Co., Ltd.},
  city={Nanjing},
  postcode={210019},
  country={China}
}

\affiliation[ippt]{
  organization={Systems Research Institute, Polish Academy of Sciences},
  city={Warsaw},
  postcode={01-447},
  country={Poland}
}

\affiliation[cityu]{
  organization={Department of Data Science, City University of HongKong},
  city={Hong Kong},
  postcode={999077},
  country={China}
}

\affiliation[spu]{
  organization={Suzhou Polytechnic University},
  city={Suzhou},
  postcode={215104},
  country={China}
}

\begin{abstract}
Interchanges are junctions where traffic merges, diverges and weaves between highways. Accurate prediction of ramp traffic flow at highway interchanges is essential for proactive traffic management involving flow control and ramp management. In practice, real-time mainline flow can be obtained through Electronic Toll Collection (ETC) systems and ramp flow can be inferred from license plate matching. However, data privacy restrictions and matching latency prevent real-time access to ramp data, leading to a “real-time blind spot”. This discrepancy dictates that while the model can utilize both mainline and ramp flow data during the training phase, the input is strictly limited to the real-time observable mainline flow during actual deployment. To address this challenge, this paper proposes a novel two-stage framework, including pre-training and prediction. In the pre-training stage, the Spatio-Temporal Decoupled Autoencoder (STDAE) leverages a proxy reconstruction task to mitigate missing ramp data issues. STDAE learns to reconstruct historical ramp flows exclusively from mainline traffic data, thereby compelling the model to capture the intrinsic spatio-temporal relationship between the mainline and ramp traffic flows. The unique decoupled architecture of STDAE, consisting of parallel spatial (SAE) and temporal (TAE) autoencoders, efficiently extracts the temporal and spatial dependency features of traffic flow. In the downstream prediction stage, the learned representations are integrated with specific forecasting models such as GWNet to enhance prediction accuracy. Comprehensive experiments conducted on three real-world interchange datasets (QiLin, DanYangXinQu, and XueBu) across multiple sampling intervals (3, 5, and 10 minutes) demonstrate that our proposed combined model, STDAEGWNET, consistently outperforms thirteen state-of-the-art baselines. Specifically, STDAEGWNET achieves the best overall average ranks of 1.00, 2.11, and 1.44 across the three respective datasets evaluated by multiple metrics. At the 3-minute sampling interval, the model attains the lowest Mean Absolute Errors (MAE) of 4.89, 5.61, and 4.58. Furthermore, the model demonstrates strong robustness against missing mainline data scenarios, achieving an average MAE reduction of approximately 2.23\% compared to baseline models without the STDAE module. Importantly, the architecture-agnostic nature enables STDAE to serve as a plug-and-play enhancement module to improve diverse forecasting pipelines. The code and datasets used in this study are publicly available at: https://github.com/ChaochaoSeu/RPaI-STDAE.
\end{abstract}




\end{frontmatter}


\section{Introduction}
\label{Introduction}
Highways are the backbone of transportation networks in urban agglomerations, providing high capacity for the movement of people and goods between origins and destinations \cite{guan2008statistical}. The highway network is composed of basic road segments, entrance and exit ramps, and system interchanges \cite{manual2000highway}. As key nodes connecting different highways, interchanges enable traffic flow to achieve continuous and rapid transitions in multiple directions through complex ramp systems. Interchange structures vary widely, including trumpet, turbine, cloverleaf, and their variants \cite{garber1999guidelines}. During peak hours, a large number of vehicles performing route changes converge at these nodes, resulting in traffic bottlenecks. These bottlenecks reduce interchange efficiency and propagate along the mainline, ultimately leading to regional congestion \cite{ma2024analysis, mu2025compliance}. Such congestion seriously undermines public travel satisfaction \cite{higgins2018all} and logistics efficiency \cite{hu2018joint}. Therefore, it is essential to monitor traffic operations at interchanges and to design advanced and effective management strategies. Accurate ramp traffic flow prediction provides the foundation for refined management \cite{ma2020statistical, liu2016model}. It helps traffic authorities analyze the mainline flow composition, identify sources and destinations, and design targeted control strategies for dynamic network management \cite{wang2022make, zhang2022adapgl, han2020hierarchical}. Ramp flow data also support traffic simulation and accurate state estimation \cite{kan2020novel}.

\begin{figure}[t]
  \centering
  \includegraphics[width=\textwidth]{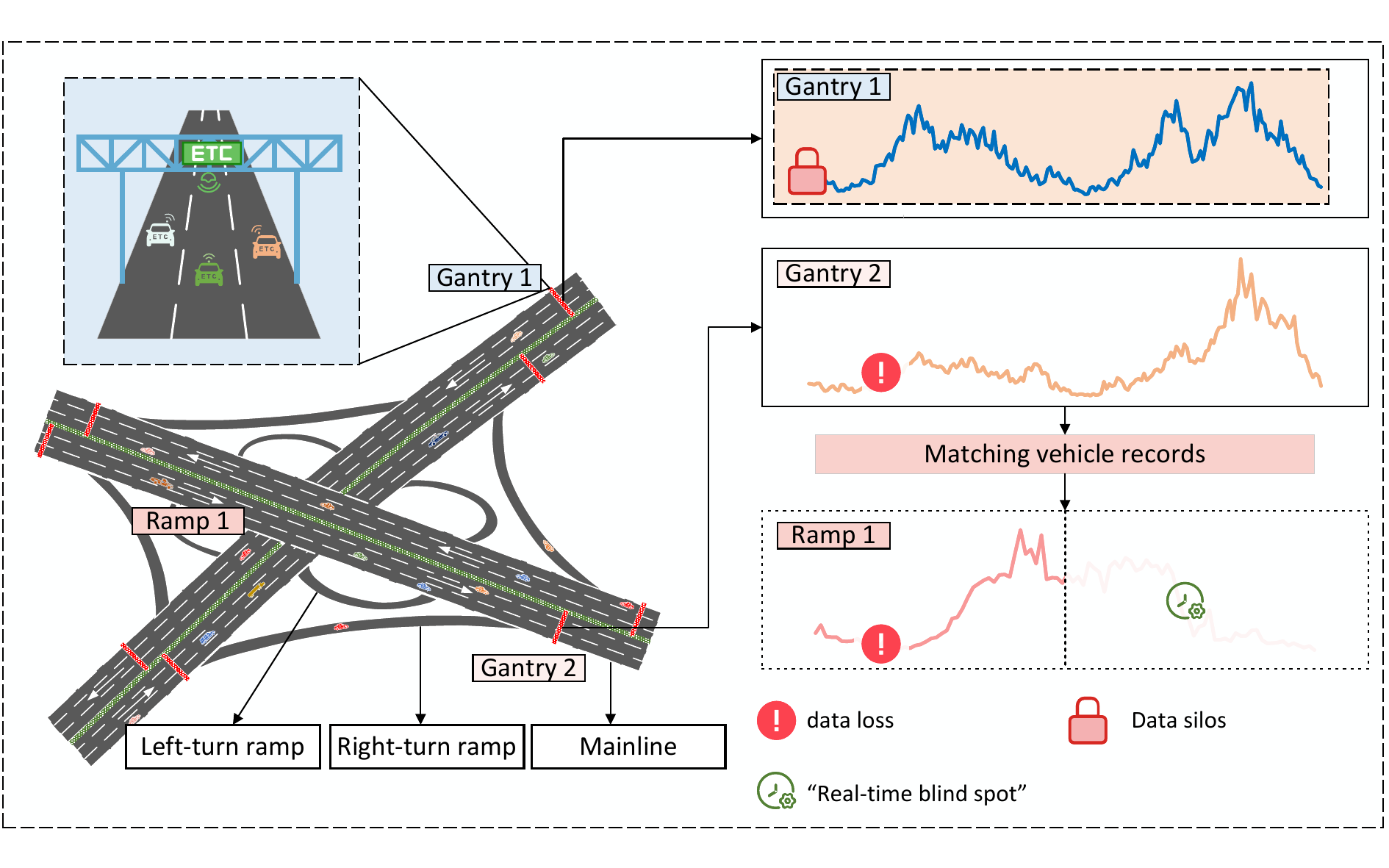}
  \caption{Illustration of the three primary challenges faced by ramp flow prediction based on mainline data.
(a) Data silos: vehicle records are fragmented among different highway operators, preventing cross-regional vehicle tracking.
(b) Data loss: sensor failures, adverse weather, or transmission errors can cause missing values even in mainline data.
(c) Real-time blind spot: real-time ramp flow data are unavailable due to privacy restrictions and system integration barriers, although historical data can be used for model training.}
  \label{fig:Description_problem}
\end{figure}

Ramp flow data are frequently unavailable because many ramps lack sensors or communication channels \cite{nohekhan2021deep}. For missing ramp detector data, imputation methods can be applied \cite{duan2016efficient}. Nevertheless, research on estimating ramp flows at locations lacking sensors remains limited. Conventional approaches primarily rely on the estimation of origin–destination (OD) matrix flows. However, the acquisition of a real-time OD matrix is extremely challenging \cite{marzano2009limits}. Machine learning (ML) methods also face significant limitations, as they require extensive datasets and are highly dependent on data quality \cite{shi2021physics, pereira2022short}. To overcome these limitations, recent studies have adopted transfer learning frameworks, utilizing mainline traffic flow to estimate ramp flows despite the lack of ramp flow data \cite{zhang2023transfer, ma2025data}.

With the widespread deployment of Electronic Toll Collection (ETC) gantry systems, high-resolution spatiotemporal traffic data on highways become increasingly accessible \cite{tsung2020visualizing, zhou2025feasibility}. Although ETC gantries are rarely installed on ramps due to high costs \cite{kan2020novel}, ramp flows can be estimated by matching vehicle records at upstream and downstream gantries \cite{zhou2025feasibility}. However, this approach encounters two primary challenges, as illustrated in \figurename~\ref{fig:Description_problem}. First, legal requirements for data privacy and the technical complexity of system integration prevent authorities from processing vehicle records in real time. Consequently, ramps experience a ``real-time blind spot'': while historical ramp flows can be obtained for model training, real-time ramp flows are unavailable as input during prediction. Second, data silos among different highway operators impede cross-regional vehicle tracking \cite{pospisil2016implementation}. Even when mainline data are available, equipment failures, adverse weather, or transmission errors often cause missing values. Therefore, developing a method that can accurately predict ramp flows based on mainline data, while also accounting for potential data loss, is a critical technical requirement for refined highway traffic management and a research topic of significant theoretical value.

Artificial Intelligence (AI) and ML methods are applied to address the challenges of ramp traffic flow prediction, offering excellent capabilities in capturing complex nonlinear relationships and spatiotemporal dependencies. From classical time series models and Support Vector Machines (SVMs) to advanced deep learning models such as Convolutional Neural Networks (CNNs) \cite{jiang2022graph}, Recurrent Neural Networks (RNNs) \cite{tedjopurnomo2020survey}, and Transformers \cite{ijcai2023p759}, these methods are widely applied to traffic prediction. For ramp flow estimation, studies use Random Forests \cite{kan2020novel} and deep learning models for hourly predictions \cite{zhang2023transfer}. Their common goal is to capture dynamic spatiotemporal dependencies in traffic data. Nevertheless, these approaches face limitations, as most require synchronous ramp data, which is infeasible under “real-time blind spot” conditions. Many also assume complete datasets, overlooking the prevalence of missing data in real-world scenarios. Transfer learning can mitigate some of these issues but remains limited for complex interchanges. It inadequately captures the spatiotemporal coupling between mainline and ramp flows, as it does not fully exploit ETC-based historical ramp data. This highlights the need for a universal model capable of handling diverse interchanges.

To address these gaps, this paper proposes the Spatio-Temporal Decoupled Autoencoder (STDAE) for ramp flow reconstruction. The framework is inspired by Masked Autoencoders (MAEs) \cite{he2022masked}, but fundamentally differs in its core objective. While classical MAE focuses on intra-modal self-reconstruction, using partial data to reconstruct missing parts of the same sequence, our approach introduces a proxy reconstruction pre-training task. This acts as a cross-regional mapping reconstruction. It operates in two stages: (1) mapping learning, where the model reconstructs historical ramp flows exclusively from observable mainline flows. The proxy reconstruction does not merely learn the internal distribution of a single traffic flow. More importantly, it forces the model to capture the complex physical and spatial mapping relationships between the mainline and the ramps, effectively bridging the "real-time blind spot"; and (2) downstream prediction, where the learned representations are transferred for accurate ramp flow forecasting. The framework requires only real-time mainline data when deployed, even if incomplete, making it highly practical. Its design not only addresses missing input data but also improves adaptability and generalization across diverse interchanges.

The contributions of this study are as follows:
\begin{itemize}
    \item \textbf{A proxy reconstruction framework for ramp flow data:} This framework reconstructs ramp flow data, which is difficult to obtain, using simpler and more easily observable mainline flow sequences. This effectively addresses the model deployment challenges caused by ramp "real-time data blind spot".
    \item \textbf{Prediction task design for missing data:} Drawing on the principles of MAE, this framework designs temporal and spatial masks before encoding to simulate missing data. This allows the model to reconstruct complete ramp data from missing mainline data, thereby improving the accuracy of the prediction model.
    \item \textbf{Comprehensive real-world validation:} The reconstruction and prediction performance is systematically evaluated on three real highway interchange datasets. A commonly used sampling interval of 5 min is selected \cite{WOS:001347142804014, shao2024exploring}, along with 3 min and 10 min, to assess the model’s adaptability to different temporal requirements. The results demonstrate the effectiveness of the proposed framework.
\end{itemize}

The rest of the paper is organized as follows. Section~\ref{Related work} reviews related work. Section~\ref{Methodology} introduces the framework and methodology. Section~\ref{Dataset and processing} describes data and preprocessing. Section~\ref{Experiments and results} presents results and discussion. Section~\ref{Conclusion} concludes the study.

\section{Related work}
\label{Related work}
\subsection{Short-term traffic prediction}
\label{Short-term traffic prediction}
Short-term traffic flow prediction is an essential component of Intelligent Transportation Systems (ITS) \cite{zhao2019t}. Accurate traffic flow forecasting provides critical decision support for traffic management authorities and helps effectively prevent and mitigate congestion \cite{liu2024spatial}. Early studies mainly relied on classical statistical methods, such as autoregressive integrated moving average (ARIMA) models \cite{hamed1995short,zhang2003time,williams2003modeling} and Kalman filter models \cite{okutani1984dynamic}. These approaches are structurally simple, easy to implement, and can provide reliable baseline predictions under stable traffic conditions. However, traffic flow is inherently nonlinear and non-stationary, forming a complex time series. Traditional statistical models struggle to capture its intrinsic stochastic fluctuations and complex spatiotemporal dependencies, resulting in limited prediction accuracy under dynamic traffic conditions \cite{9790148}.

Beyond traffic engineering, modeling complex nonlinear spatiotemporal patterns and handling missing data are pervasive challenges across many scientific disciplines. To address these, the broader machine learning literature offers diverse techniques. For instance, Gaussian process regression is widely used in environmental and energy forecasting to model nonlinear time series and quantify uncertainty \cite{sheng2017short,fang2018novel}. Similarly, advanced graph and probabilistic models excel at capturing complex dependencies in multivariable systems \cite{sun2022deep,tang2019long}, while ensemble methods aggregate diverse algorithms to improve generalization and mitigate noisy data \cite{goehry2019aggregation}. Driven by these cross-disciplinary advancements, intelligent transportation researchers increasingly adopt machine learning to capture nonlinear traffic flow dynamics \cite{tedjopurnomo2020survey}.

Shallow methods, such as SVM \cite{chen2015forecasting}, K-Nearest Neighbors (KNN) \cite{zhang2009short}, and hybrid models \cite{li2023qpso}, can flexibly capture nonlinear relationships without requiring strict stationarity assumptions and exhibit strong adaptability to noise and local traffic flow trends. Consequently, these methods achieve higher prediction accuracy under complex traffic conditions. With the rapid development of research, deep learning has gradually become the mainstream approach \cite{tedjopurnomo2020survey,9352246,jiang2022graph}. In particular, RNNs and their variants, especially Long Short-Term Memory (LSTM) networks \cite{hu2025novel,dong2025real} and Gated Recurrent Units (GRU) \cite{dai2019short}, demonstrate outstanding performance in capturing temporal dependencies. Meanwhile, CNNs are employed to extract spatial topological features and combined with RNNs to form hybrid models \cite{jiang2018geospatial,mendez2023long}. To further characterize the complex topological structure of traffic networks, researchers propose spatiotemporal prediction models based on Graph Neural Networks (STGNN) \cite{wu2020comprehensive,WOS:000764175403107,zhao2019t,10151673,HE2025127564}, enhancing the model's ability to capture spatiotemporal dependencies. Aattention mechanisms and Transformers are also introduced for traffic prediction \cite{liu2024itransformer,jin2025ada,dong2025transformer}, enabling dynamic focus on critical time periods and key road segments, thereby improving the robustness and accuracy of predictions.

It is noteworthy that the majority of these studies are applied to urban arterials or freeway mainlines where data collection is relatively comprehensive. The development and validation of these methods heavily depend on accessible mainline traffic flow data. In contrast, ramp traffic flows, which are often sparsely instrumented and difficult to observe directly, receive relatively limited attention in the literature, representing a significant gap in current research.
\subsection{Ramp flow prediction}
\label{Ramp flow prediction}
Highway ramp traffic flow prediction is a specialized branch within the field of traffic flow forecasting. The core task is to estimate the number of vehicles that will merge and diverge via ramps over a given future time interval \cite{zhang2023transfer}. As noted in Section~\ref{Introduction}, obtaining ramp flow data is challenging. Thus, existing research has primarily focused on methods for estimating flows at ramps without sensors. The first three rows of Table~\ref{table1} provide a comprehensive comparison of different types of ramp traffic flow prediction models. Traditional approaches mainly rely on OD matrices to allocate link flows. However, the real-time updating of OD matrices is extremely difficult (RTC: \ding{55}), making these methods insufficient for dynamic traffic management \cite{marzano2009limits}. Some studies combine traffic flow models with the OD matrix, but these methods typically require complete mainline and ramp data (DI: M-C, R-C) and are computationally intensive \cite{chang2002optimization}. They also struggle to cope with missing data.

Data-driven methods have become mainstream due to their powerful feature capture capabilities. As shown in Table~\ref{table1}, this strong reliance on auxiliary information limits the model's scenario generalization ability (SGA). For instance, Nohekhan et al. \cite{nohekhan2021deep} proposed a model that requires static or semi-static inputs such as the annual average daily traffic (AADT), road classification, and free-flow speed. Similarly, Kan et al. \cite{kan2020novel} relied on ramp capacity and peak demand parameters derived from the Highway Capacity Manual (HCM) \cite{manual2000highway}. The model requires the integrity of the mainline data (DI: M-C), making it susceptible to data loss.

\renewcommand{\arraystretch}{1.1}
\begin{table}[]
\centering
\caption{Comparative analysis of different traffic prediction models across multiple evaluation metrics}
\label{table1}
\footnotesize
\adjustbox{width=\textwidth,center}{
\begin{tabular}{ccccccccccclc}
\toprule
\multirow{2}{*}{Model\textsuperscript{1}} & \multirow{2}{*}{Target\textsuperscript{2}} & \multicolumn{3}{c}{Data Conditions\textsuperscript{3}} & \multicolumn{3}{c}{Modeling Paradigm\textsuperscript{4}} & \multicolumn{3}{c}{Model Evaluation\textsuperscript{5}} & \multicolumn{1}{c}{\multirow{2}{*}{Reference}} &  \\ \cline{3-11}
 &  & IDS & DI & DT & Type & TDM & SDM & VM & RTC & SGA & \multicolumn{1}{c}{} &  \\ \midrule
OD & RFP & M-S, R-S & M-C, R-C & RT & T & $\checkmark$ & $\times$ & SB & \ding{55} & L & \cite{chang2002optimization} &  \\
RF/GBM & RFP & M-S & M-C, R-M & RT & ML & $\checkmark$ & $\times$ & RW & $\checkmark$ & M & \cite{kan2020novel} &  \\
DDA+MT & RFP & M-S & M-C, R-M & RT & TL & $\checkmark$ & $\times$ & RW & $\checkmark$ & M & \cite{zhang2023transfer} &  \\
TST & MFP & M-S & R & Pre-N, Test-RT & Pre & $\checkmark$ & $\times$ & RW & $\checkmark$ & H & \cite{zerveas2021transformer} &  \\
STEP & NFP & M-LS & R & Pre-N, Test-RT & Pre & $\checkmark$ & $\checkmark$ & RW & $\checkmark$ & H & \cite{shao2022pre} &  \\
DGCN-PTL & NFP & M-LS & R & Pre-N, Test-RT & Pre & $\checkmark$ & $\checkmark$ & RW & $\checkmark$ & H & \cite{chi2025dynamic} &  \\
STDMAE & NFP & M-LS & R & Pre-N, Test-RT & Pre & $\checkmark$ & $\checkmark$ & RW & $\checkmark$ & H & \cite{WOS:001347142804014} &  \\
TS-MAE & MFP & M-S & R & Pre-N, Test-RT & Pre & $\checkmark$ & $\times$ & RW & $\checkmark$ & H & \cite{liu2025ts} &  \\
PreSTNet & NFP & M-LS & R & Pre-N, Test-RT & Pre & $\checkmark$ & $\checkmark$ & RW & $\checkmark$ & H & \cite{fang2024prestnet} &  \\
JointSTNet & NFP & M-S & R & Pre-N, Test-RT & Pre & $\checkmark$ & $\checkmark$ & RW & $\checkmark$ & H & \cite{cai2024jointstnet} &  \\
\textbf{Ours} & RFP & M-LS, R-LS & R & Pre-N, Test-RT & Pre & $\checkmark$ & $\checkmark$ & RW & $\checkmark$ & H & -- &  \\ \bottomrule
\end{tabular}
}
\begin{flushleft}
\footnotesize
\textbf{Note:} 
\textsuperscript{1} Model: Name of the evaluated traffic flow prediction model. \\
\textsuperscript{2} Target: Application domain, including Ramp Flow Prediction (RFP), Mainline Flow Prediction (MFP), and Network-Wide Flow Prediction (NFP). \\
\textsuperscript{3} Data Conditions: Dataset assessment including Input Data Source (IDS), Data Integrity (DI), and Data Timeliness (DT); M-S/M-LS: Mainline Short/Long–Short series; R-S/R-LS: Ramp Short/Long–Short series; M-C/R-C: Mainline/Ramp Complete data; R-M: Ramp Missing data; R: Robust to missing data. \\
\textsuperscript{4} Modeling Paradigm: Includes model type (Type), Temporal Dependency Modeling (TDM), and Spatial Dependency Modeling (SDM); T: Traffic Flow Model, ML: Machine Learning, TL: Transfer Learning, Pre: Pre-training Model. \\
\textsuperscript{5} Model Evaluation: Includes Validation Method (VM), Real-Time Capability (RTC), and Scenario Generalization Ability (SGA); RW: Real-World, SB: Simulation/Benchmark; L: Low, M: Medium, H: High.
\end{flushleft}

\end{table}

To alleviate the reliance on training data, transfer learning is introduced into ramp flow estimation. The core idea is to transfer knowledge learned from a data-rich source domain (e.g., mainline) to a data-scarce target domain (e.g., ramps) \cite{zhuang2020comprehensive}, providing a feasible solution for ramps without historical data. Although traffic states differ across highway segments, they exhibit inherent correlations, creating a natural scenario for the application of transfer learning \cite{zhuang2020comprehensive}. In ramp flow prediction, transfer learning is mainly applied in two ways: feature-representation transfer, which reduces the discrepancy in marginal distributions between the source and target domains, and parameter transfer, which mitigates differences in conditional distributions. Combining these strategies can significantly improve prediction accuracy and generalization capability in the target domain \cite{zhang2023transfer}. Transfer learning methods encounter new challenges when applied to structurally complex interchange scenarios. First, most prior studies focus on ramps with simple structures. Interchanges, however, have diverse layouts and complex ramp systems \cite{garber1999guidelines}. This makes it difficult to directly apply existing models. As shown in Table~\ref{table1}, they exhibit limited scene generalization ability (SGA) in interchange scenarios. Second, the data environment has evolved. With the deployment of ETC gantries, ramp flow data can now be obtained through vehicle path matching. However, ramp flow prediction faces the "real-time blind spot" problem in Section~\ref{Introduction}. Transfer learning methods usually rely only on mainline data and ignore the valuable historical ramp data provided by ETC systems. This limits the ability of models to capture the complex spatiotemporal dependencies between mainline and ramp flows. Therefore, the key challenge is to use limited historical data effectively while handling incomplete mainline inputs. This is essential for developing accurate predictive models for ramps in complex interchanges.

\subsection{Pre-training framework for traffic prediction}
Inspired by successful practices in the field of natural language processing (NLP), pre-training has become a key paradigm for improving model performance \cite{liu2023pre}. The Transformer architecture, due to its strong ability to handle long sequences, forms the core of such pretrained models \cite{vaswani2017attention}. In this context, researchers extend the idea of pre-training to time series modeling. Through self-supervised learning, models extract effective feature representations from large-scale historical data. This enhances the performance of downstream prediction tasks \cite{shao2022pre,chi2025dynamic,WOS:001347142804014}. Among these methods, MAE \cite{he2022masked} has become a mainstream pre-training strategy. The core task of MAE is to mask part of the observed data and reconstruct it using the remaining visible data. This allows the model to learn deep relationships within the data. Such a property makes MAE naturally suitable for handling missing data, providing a new approach to improving the robustness of traffic prediction models.

Existing pre-training frameworks explore various model structures and masking strategies. Rows 4 to 10 of Table~\ref{table1} present a comparison of different pre-training models for traffic flow prediction. TST \cite{zerveas2021transformer} uses a masked Transformer architecture to learn representations of multivariate time series. Other models, including STEP \cite{shao2022pre}, DGCN-PTL \cite{chi2025dynamic}, and STDMAE \cite{WOS:001347142804014}, further integrate the intermediate representations generated by MAE into downstream predictors to improve performance. Notably, STDMAE proposes a spatial-temporal decoupled masking mechanism. It performs masking and reconstruction separately along the temporal and spatial dimensions. This allows the model to better capture spatiotemporal heterogeneity. To model temporal dependencies more precisely, TS-MAE \cite{liu2025ts} constructs a masked autoencoder based on neural ordinary differential equations (Neural ODEs). This enables continuous modeling of evolution patterns. TS-MAE also uses a stepwise masking strategy to enhance learning robustness. In addition, some studies incorporate graph structures into pre-training. PesNet \cite{fang2024prestnet} uses graph convolution operators to recover masked data. JointSTNet \cite{cai2024jointstnet} introduces a spatial graph capsule module and a temporal gating module. It also combines an adaptive masking strategy to improve the extraction and learning of complex spatiotemporal relationships. These models exhibit strong real-time capability and scene generalization ability. Their main difference lies in the duration of the input data.

Although these studies (such as STDMAE \cite{WOS:001347142804014}) demonstrate the advantages of pre-training in capturing long-term and diverse spatiotemporal features, their pre-training tasks fundamentally rely on self-reconstruction. That is, they reconstruct missing parts of the same source traffic flow to learn its internal data distribution. However, this intra-modal approach cannot directly solve the problem in interchange scenarios. In these cases, data from one spatial domain (mainline traffic) must be used to predict another unobservable domain (ramp traffic). To address this, we propose a proxy reconstruction pre-training task. Unlike classical MAE, our model leverages mainline traffic flow data to reconstruct ramp traffic flows. Through this cross-regional mapping reconstruction, the model transcends learning simple historical patterns and instead learns the intrinsic physical and spatiotemporal mapping relationships between mainline and ramp flows. In this framework, the model can learn more intrinsic patterns between mainline and ramp flows. Moreover, masking can be applied to the mainline inputs. This simulates real-world scenarios with partial missing data and enhances the robustness of the prediction model.

In summary, although ramp flow prediction makes great progress, current models still have some limitations. To compare the strengths and weaknesses of different approaches more clearly, this study reviews ramp flow prediction methods together with recent pre-training frameworks. Traditional approaches mainly use traffic flow models and dynamic OD estimation. They can predict ramp flows under certain conditions. However, they require complete and real-time data, and the computation cost is high. Machine learning methods reduce the impact of missing data by using additional static or semi-static features. But they rely heavily on auxiliary information, which limits scalability. Transfer learning methods transfer knowledge from mainline flows to ramps. This improves prediction accuracy, but they remain insufficient in complex interchange scenarios. Pre-training approaches take a different path. They learn representations from historical data through self-supervised learning. These methods show strong robustness and generalization. They also perform well under missing data conditions.

\section{Methodology}
\label{Methodology}
\begin{figure}[t]
  \centering
  \includegraphics[width=\textwidth]{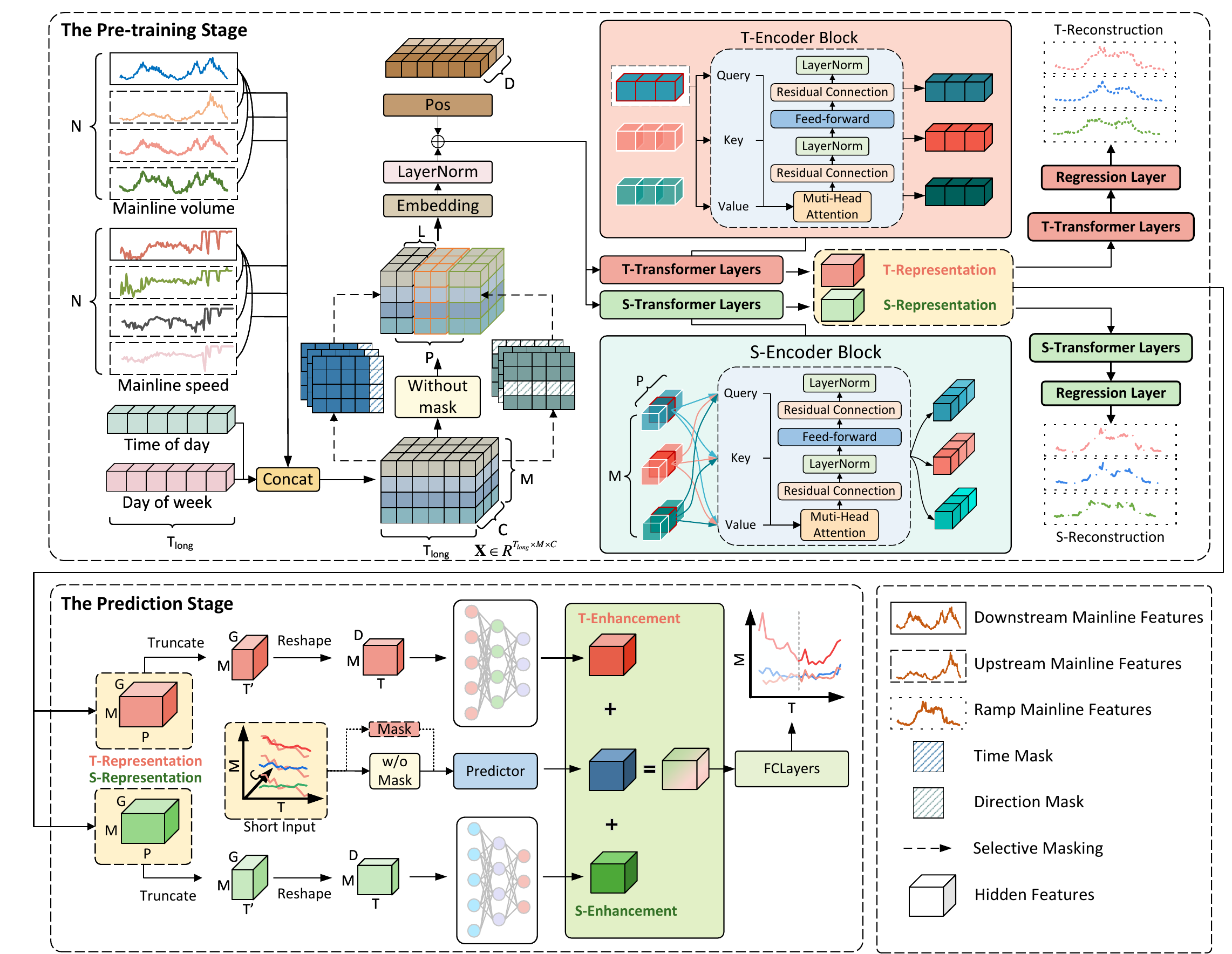}
  \caption{STDAE Pre-training and Prediction Framework}
  \label{fig:stdae}
\end{figure}

This section describes the proposed model in detail. The overall framework of the model is illustrated in \figurename~\ref{fig:stdae}. The STDAE model consists of two components: a pre-training model and a downstream prediction model. In the pre-training stage, an STDAE structure is designed to more efficiently capture the spatio-temporal dependencies of traffic data. This structure comprises two mutually independent modules: a Temporal AutoEncoder (TAE) and a Spatial AutoEncoder (SAE). Both modules share the same architecture but focus separately on reconstructing features in the temporal and spatial dimensions, respectively, to generate temporal and spatial representations of ramp traffic flow states. In the downstream prediction stage, these representations are processed through a multi-layer perceptron (MLP) module and fused with the hidden layer of the prediction model to obtain the final forecast of ramp traffic flow.

\subsection{Problem Statement}
\label{ProblemStatement}

In this study, the short-term prediction of ramp traffic flow at interchanges is defined as a temporal prediction problem. Consider a hub interchange with two mainlines crossing, where the mainlines consist of $N = 8$ directions, including four upstream and four downstream directions for the two intersecting lines. There are $M = 12$ traffic movements (ramps), with each ramp connecting the upstream and downstream mainline flows from two corresponding directions.

$\mathbf{V}_{t-(T-1):t} \in \mathbb{R}^{T \times N \times F}$ denotes the historical mainline traffic flow feature time series for $T$ consecutive time steps, where $F$ represents the number of features per observation point. The objective is to predict the ramp traffic flow $\mathbf{Y}_{t+1:t+S} \in \mathbb{R}^{S \times M \times I}$ for the next $S$ time steps, given the ramp traffic graph $\mathcal{G}$:

\begin{equation}
[\hat{\mathbf{Y}}_{t+1}, \ldots, \hat{\mathbf{Y}}_{t+S}] = F(\mathbf{V}_{t-(T-1):t}, \mathcal{G}),
\end{equation}
where $\hat{\mathbf{Y}}_j \in \mathbb{R}^{M \times I}$ represents the predicted ramp traffic flow at the $j$th time step. In the dataset used in this study, $I = 1$, corresponding to traffic volume measurements.

To capture the dependencies among ramps, a graph $\mathcal{G} = (\mathcal{V}, \mathcal{E})$ is constructed, where each node corresponds to a ramp and each edge denotes spatial or functional correlations among ramps. The graph has $M$ nodes corresponding to the $M$ ramps, and its adjacency matrix $\mathbf{A} \in \mathbb{R}^{M \times M}$ encodes the pairwise relationships among ramps:

\begin{equation}
\mathbf{A}_{i,j} =
\begin{cases}
1, & \text{if ramp $i$ and ramp $j$ are connected or correlated}, \\
0, & \text{otherwise}.
\end{cases}
\end{equation}

In the ablation experiments, we evaluate multiple graph construction strategies to assess the sensitivity of the model to prior topology. Experimental results demonstrate that initializing the graph by connecting physically proximate ramps provides stable and accurate performance. Therefore, this configuration is selected as the final prior adjacency structure.

\subsection{The pre-training model}
In the pre-training stage, we employ a Transformer-based autoencoder rather than a graph-structured model. The rationale behind this design lies in the objective of proxy reconstruction, which aims to capture long-range and global dependencies between mainline and ramp traffic flows. Since there is no explicit or predefined graph structure that directly characterizes the heterogeneous mapping between these components, adopting a GNN-based pre-training strategy would impose localized topological constraints that may limit the discovery of cross-regional correlations. In contrast, the self-attention mechanism of Transformers is free from predefined adjacency priors and is inherently capable of modeling non-local and dynamic interactions across the entire interchange. Instead of using a conventional self-reconstruction autoencoding scheme that reconstructs the input itself, we design a proxy task in which ramp traffic flow sequences are reconstructed from long historical time series of upstream and downstream mainline flows. To effectively model the complex relationships between heterogeneous traffic streams, we introduce a spatiotemporal decoupling strategy that separates spatial dependencies from temporal dependencies and learns their representations independently. This design enables more precise modeling of intricate spatiotemporal dynamics. Furthermore, an optional masking module is incorporated to improve robustness under partial mainline data missing scenarios.

\subsubsection{Input Feature Encoding}
\label{Input Feature Encoding}

The input feature encoding process of the model consists of two core stages: Feature Fusion and Feature Encoding. The goal is to transform raw spatiotemporal data into informative embeddings suitable for STDAE.

The objective of feature fusion is to integrate mainline traffic data to construct a comprehensive feature representation for each ramp. The raw input consists of sequential mainline traffic data, including traffic flow, speed, day of the week, time of day, and other attributes. This data is represented as $\mathbf{V}_{t-({T}_{long}-1):t} \in \mathbb{R}^{{T}_{long} \times N \times F}$, where $T_{\text{long}}$ denotes the length of the input time series. Next, these mainline features are reorganized and aggregated to construct ramp features. For the $m$-th ramp, $m\in \{1, \dots, M\}$, let $U(m)$ and $D(m)$ denote the indices of its upstream and downstream mainline detectors, respectively. At any time step $\tau$, the feature vector of ramp $m$ is formed by concatenation:

\begin{equation}
\mathbf{x}_{\tau, m} = \text{Concat}_{n \in U(m) \cup D(m)} \big( \mathbf{v}_{\tau, n} \big),
\end{equation}
where $\mathbf{v}_{\tau, n} \in \mathbb{R}^{F}$ represents the feature vector of mainline detector $n$ at time $\tau$.  

By applying this procedure to all $M$ ramps and $T_{\text{long}}$ time steps, the fused feature tensor $\mathbf{X} \in \mathbb{R}^{T_{\text{long}} \times M \times C}$ is obtained.
where $C = (\lvert U(m)\rvert + \lvert D(m)\rvert) \cdot F $ depends on the number of upstream and downstream connections. This step is applied during both pre-training and prediction stages to ensure consistent input representation.

The feature encoding stage transforms the fused tensor $\mathbf{X}$ into the final input embedding $\mathbf{E}$, consisting of masking, patching \& embedding, and positional encoding.

To simulate missing mainline data, spatial and temporal masks, denoted as $\mathbf{M}_{\mathrm{s}}$ and $\mathbf{M}_{\mathrm{t}}$, are defined as binary tensors of the same shape as $\mathbf{X}$, where 1 indicates retained data and 0 indicates masked data. The masking operation is applied as:
\begin{equation}
\mathbf{X}' = \mathbf{X} \odot \mathbf{M}_{\mathrm{s}} \odot \mathbf{M}_{\mathrm{t}},
\end{equation}
where $\odot$ denotes the Hadamard product.  

For the spatial mask:
\begin{equation}
(\mathbf{M}_{\mathrm{s}})_{\tau, m, c} = 
\begin{cases}
0, & \text{if node } m \text{ and dimension } c \text{ match the rule} \\
1, & \text{otherwise}
\end{cases}
\end{equation}

For the temporal mask:
\begin{equation}
(\mathbf{M}_{\mathrm{t}})_{\tau, m, c} = 
\begin{cases}
0, & \tau \in \text{mask time list} \\
1, & \text{otherwise}
\end{cases}
\end{equation}

If the data is complete, no masking is applied, i.e., $\mathbf{X}' = \mathbf{X}$.

To reduce computational cost for long sequences $T_{\text{long}}$, the time dimension is divided into $P$ non-overlapping patches, each of length $L = T_{\text{long}} / P$. Let the reshaped input for the convolutional patch embedding layer be $\mathbf{X}_{\text{conv}} \in \mathbb{R}^{M \times C \times T_{\text{long}}}$.
The 2D convolution kernel $\mathbf{W}_{\text{conv}} \in \mathbb{R}^{D \times C \times L \times 1}$ with bias $\mathbf{b}_{\text{conv}} \in \mathbb{R}^{D}$ is applied:
\begin{equation}
\mathbf{E}_p = \text{Conv2D}(\mathbf{X}_{\text{conv}}) \in \mathbb{R}^{M \times P \times D}.
\end{equation}

Before adding positional information, the embedding $\mathbf{E}_p$ is normalized along the feature dimension:
\begin{equation}
\mathbf{E}_{\mathrm{norm}} = \text{LayerNorm}(\mathbf{E}_p),
\end{equation}
where for any vector $\mathbf{z} \in \mathbb{R}^D$:
\begin{equation}
\text{LayerNorm}(\mathbf{z}) = \frac{\mathbf{z} - \mu}{\sqrt{\sigma^2 + \epsilon}} \odot \gamma + \beta,
\end{equation}
with $\mu$ and $\sigma^2$ being the mean and variance of $\mathbf{z}$, $\gamma, \beta \in \mathbb{R}^D$ learnable parameters, and $\epsilon$ a small constant.

To encode spatial and temporal positions, a 2D sinusoidal positional encoding is defined \cite{wang2021translating}:
\begin{equation}
(\mathbf{E}_{\mathrm{pos}})_{m,p,d} =
\begin{cases}
\sin\left(\dfrac{m}{10000^{2i/D_{\text{half}}}}\right), & d = 2i, \; d < D_{\text{half}}, \\[8pt]
\cos\left(\dfrac{m}{10000^{2i/D_{\text{half}}}}\right), & d = 2i+1, \; d < D_{\text{half}}, \\[8pt]
\sin\left(\dfrac{p}{10000^{2i/D_{\text{half}}}}\right), & d = D_{\text{half}} + 2i, \\[8pt]
\cos\left(\dfrac{p}{10000^{2i/D_{\text{half}}}}\right), & d = D_{\text{half}} + 2i + 1,
\end{cases}
\end{equation}
where $D_{\text{half}} = D/2$ and $i$ indexes the feature dimensions.

Finally, the normalized embedding is combined with positional encoding to form the final input:
\begin{equation}
\mathbf{E} = \mathbf{E}_{\mathrm{norm}} + \mathbf{E}_{\mathrm{pos}} \in \mathbb{R}^{M \times P \times D}.
\end{equation}

This tensor $\mathbf{E}$ contains both the semantic information of the original data and the spatiotemporal structure, serving as input to subsequent Transformer-based models.

\subsubsection{STDAE}
Spatiotemporal decoupling is the core design principle of STDAE for modeling interactions between mainline and ramp traffic flows. Unlike coupled architectures such as STGCN \cite{WOS:000764175403107} and DCRNN \cite{li2017diffusion} that jointly learn spatial–temporal dependencies, our framework explicitly separates these dimensions. This design is physically motivated: spatial correlations are primarily determined by road topology, whereas temporal dynamics arise from periodic demand patterns and traffic wave propagation. Decoupling enables more precise representation learning, alleviates feature interference and over-smoothing, and reduces computational complexity compared with high-dimensional joint attention mechanisms \cite{WOS:001347142804014}. Specifically, a stack of Transformer encoder layers is employed to extract heterogeneous spatial features by modeling interactions between mainline and ramp flows, thereby capturing latent spatial dependencies among ramps. The resulting embedding $\mathbf{E}$, derived from mainline traffic features, is then fed into two parallel modules with identical architectures but distinct learning objectives.

\textbf{Spatial Reconstruction Module (SAE).} 
This module extracts the spatial heterogeneity features of ramp and mainline traffic flows through a stack of Transformer encoder layers. By leveraging the interaction between mainline and ramp flows, the module is able to capture the underlying spatial dependencies across different ramps. 

To enable the model to focus on the spatial dimension, the input embedding $\mathbf{E}$ is first transposed along the spatial ($M$) and temporal ($P$) dimensions. The transposed tensor $\mathbf{E}^T \in \mathbb{R}^{M \times P \times D}$ is then fed into the spatial encoder. Within each encoder layer, the spatial self-attention is computed among all ramps at each time step, formulated as:
\begin{equation}
\text{Attn}^{(S)}(\mathbf{Q}_S, \mathbf{K}_S, \mathbf{V}_S) = \text{Softmax}\left( \frac{\mathbf{Q}_S \mathbf{K}_S^{\top}}{\sqrt{d_k}} \right)\mathbf{V}_S,
\end{equation}
where $\mathbf{Q}_S = \mathbf{E}^T \mathbf{W}_Q^{(S)}$, $\mathbf{K}_S = \mathbf{E}^T \mathbf{W}_K^{(S)}$, and $\mathbf{V}_S = \mathbf{E}^T \mathbf{W}_V^{(S)}$ denote the query, key, and value matrices, and respectively.  
This attention mechanism allows each ramp to attend to all other ramps (and the mainline) at the same time step, thereby modeling the spatial correlations of traffic states.  

After encoding and layer normalization, the result is transposed back to the original dimension order to form the spatial representation $\mathbf{H}^{(S)} \in \mathbb{R}^{M \times P \times D}$:
\begin{equation}
\mathbf{H}^{(S)} = \text{LayerNorm}\big(\text{EncoderS}(\mathbf{E}^T)^{T}\big).
\end{equation}

The spatial representation $\mathbf{H}^{(S)}$ is first passed through a linear mapping layer and then transposed again to match the input requirements of the spatial decoder. The decoder reconstructs the spatial sequences through the same type of attention operation:
\begin{equation}
\text{Attn}^{(S)}_{\text{dec}}(\mathbf{Q}'_S, \mathbf{K}'_S, \mathbf{V}'_S) = \text{Softmax}\left( \frac{\mathbf{Q}'_S \mathbf{K}'^{\top}_S}{\sqrt{d_k}} \right)\mathbf{V}'_S,
\end{equation}
followed by normalization and an output projection to obtain the reconstructed spatial sequence:
\begin{equation}
\hat{\mathbf{Y}}^{(S)} = \Big[\text{OutputLayer}\big(\text{LayerNorm}(\text{DecoderS}(\text{Linear}(\mathbf{H}^{(S)})^T))\big)\Big]^T.
\end{equation}

\textbf{Temporal Reconstruction Module (TAE).} 
This module shares the same architecture as SAE but focuses on capturing long-range temporal dependencies rather than spatial ones.

In the temporal domain, the input embedding $\mathbf{E} \in \mathbb{R}^{M \times P \times D}$ is directly fed into the temporal encoder. 
Here, the temporal self-attention operates along the time dimension, enabling each time step to attend to all other time steps for the same ramp and mainline flow:
\begin{equation}
\text{Attn}^{(T)}(\mathbf{Q}_T, \mathbf{K}_T, \mathbf{V}_T) = \text{Softmax}\left( \frac{\mathbf{Q}_T \mathbf{K}_T^{\top}}{\sqrt{d_k}} \right)\mathbf{V}_T,
\end{equation}
where $\mathbf{Q}_T = \mathbf{E}\mathbf{W}_Q^{(T)}$, $\mathbf{K}_T = \mathbf{E}\mathbf{W}_K^{(T)}$, and $\mathbf{V}_T = \mathbf{E}\mathbf{W}_V^{(T)}$.  
This formulation allows the model to capture long-range temporal dependencies, including recurrent congestion patterns and propagation delays between mainline and ramps.  
After applying temporal encoding and layer normalization, the temporally encoded representation is obtained as:
\begin{equation}
\mathbf{H}^{(T)} = \text{LayerNorm}(\text{EncoderT}(\mathbf{E})).
\end{equation}

The temporal representation $\mathbf{H}^{(T)}$ is linearly mapped and directly input into the temporal decoder without transposition. The decoder applies the same temporal attention mechanism:
\begin{equation}
\text{Attn}^{(T)}_{\text{dec}}(\mathbf{Q}'_T, \mathbf{K}'_T, \mathbf{V}'_T) = \text{Softmax}\left( \frac{\mathbf{Q}'_T \mathbf{K}'^{\top}_T}{\sqrt{d_k}} \right)\mathbf{V}'_T,
\end{equation}
followed by layer normalization and the output projection:
\begin{equation}
\hat{\mathbf{Y}}^{(T)} = \text{OutputLayer}\big(\text{LayerNorm}(\text{DecoderT}(\text{Linear}(\mathbf{H}^{(T)})))\big).
\end{equation}

The main difference between SAE and TAE lies in the dimension along which attention is computed. SAE transposes the embedding tensor to allow attention across ramps and mainline flows at a fixed time step, thereby modeling \emph{spatial dependencies}. In contrast, TAE preserves the original dimension order to perform attention across time steps, thereby modeling \emph{temporal dependencies}.  
Formally, SAE focuses on:
\begin{equation}
\mathbf{A}^{(S)} = \text{Softmax}\left( \frac{\mathbf{Q}_S \mathbf{K}_S^{\top}}{\sqrt{d_k}} \right) \quad \text{where } \mathbf{A}^{(S)} \in \mathbb{R}^{M \times M},
\end{equation}
while TAE focuses on:
\begin{equation}
\mathbf{A}^{(T)} = \text{Softmax}\left( \frac{\mathbf{Q}_T \mathbf{K}_T^{\top}}{\sqrt{d_k}} \right) \quad \text{where } \mathbf{A}^{(T)} \in \mathbb{R}^{P \times P}.
\end{equation}

Both modules are optimized by computing reconstruction losses between $\hat{\mathbf{Y}}^{(S)}$ and $\hat{\mathbf{Y}}^{(T)}$ and the true ramp sequences. 
The SAE captures spatial heterogeneity among ramps, while the TAE models long-range temporal patterns. 
By reconstructing ramp traffic based on mainline flow features, the model implicitly learns the spatiotemporal dependencies between ramps and the mainline. 
The mainline acts as the traffic backbone, influencing ramp inflows and outflows and reflecting traffic propagation and interaction. 
Consequently, the reconstructed ramp sequences encode both spatial correlations and temporal dynamics, providing complementary spatiotemporal representations for downstream ramp traffic prediction tasks.

\subsection{Downstream prediction model}
In the downstream prediction model, it is possible to seamlessly integrate the pretrained model STDAE. This operation is accomplished by taking the spatial and temporal representations generated by STDAE and adding them to the hidden representation of the predictor. 

Specifically, a long sequence of $T_{\text{long}}$ time steps is first fed into the pretrained spatial and temporal encoders to generate the corresponding spatial representation $H^{(S)}$ and temporal representation $H^{(T)}$. Subsequently, the commonly used short-term input sequence $X_{t-(T-1):t}$ is fed into the downstream prediction model to obtain the hidden representation $H^{(F)} \in \mathbb{R}^{T \times M \times D}$, where $D$ is the dimensionality of the hidden representation. During the input stage, the same masking strategy as in the pre-training stage is applied to ensure consistency. 

To align with the hidden representation \( \mathbf{H}^{(F)} \), the last \( T' \) patches from \( \mathbf{H}^{(S)} \) and \( \mathbf{H}^{(T)} \) are extracted and reshaped into tensors \( \mathbf{H}'^{(S)} \in \mathbb{R}^{T \times M \times D} \) and \( \mathbf{H}'^{(T)} \in \mathbb{R}^{T \times M \times D} \), respectively.
 These representations are then projected to the target dimension \(D'\) using two independent two-layer multilayer perceptrons.

The final augmented representation \( \mathbf{H}^{(\text{Aug})} \in \mathbb{R}^{T \times M \times D} \) is obtained by summing the projected spatial and temporal representations with the downstream hidden representation, as follows:
\begin{equation}
\mathbf{H}^{(\text{Aug})} = \text{MLP}(\mathbf{H}'^{(S)}) + \text{MLP}(\mathbf{H}'^{(T)}) + \mathbf{H}^{(F)}.
\end{equation}

Up to this point, the augmented representation \( \mathbf{H}^{(\text{Aug})} \) incorporates the short-term representation learned by the downstream prediction model itself and the long temporal dependency features provided by STDAE, thus effectively compensating for the limited ability of traditional models to capture long-term dependencies and significantly improving the overall performance and generalization of downstream spatio-temporal prediction tasks.

In this study, the downstream forecasting model adopts GWNet \cite{ijcai2019p264}. Its integration of adaptive graph convolutions and dilated causal convolutions makes it a widely recognized and powerful baseline in the spatiotemporal forecasting domain. This makes it an ideal backbone to rigorously test the effectiveness of the auxiliary features generated by STDAE. As demonstrated in Section~\ref{Ablation on downstream predictors}, STDAE successfully improves various diverse predictors, with GWNet yielding the most competitive overall performance. STDAE learns structured global representations by separately modeling spatial and temporal features, without requiring additional masking or reconstruction steps. By integrating the STDAE representations into the hidden states as auxiliary information, the model can better capture long-range dependencies, enabling more accurate and stable predictions in complex spatiotemporal tasks.

\section{Dataset and processing}
\label{Dataset and processing}
This section provides a detailed description of the entire process of data collection, cleaning, and feature engineering used in this study. The overall workflow is illustrated in \figurename~\ref{fig:dataset}. The purpose of this process is to transform raw ETC transaction records into time series data of mainline and ramp traffic states that can be utilized by the model.

\subsection{Define the scope}
\label{Define the scope}

\begin{figure}[t]
  \centering
  \includegraphics[width=\textwidth]{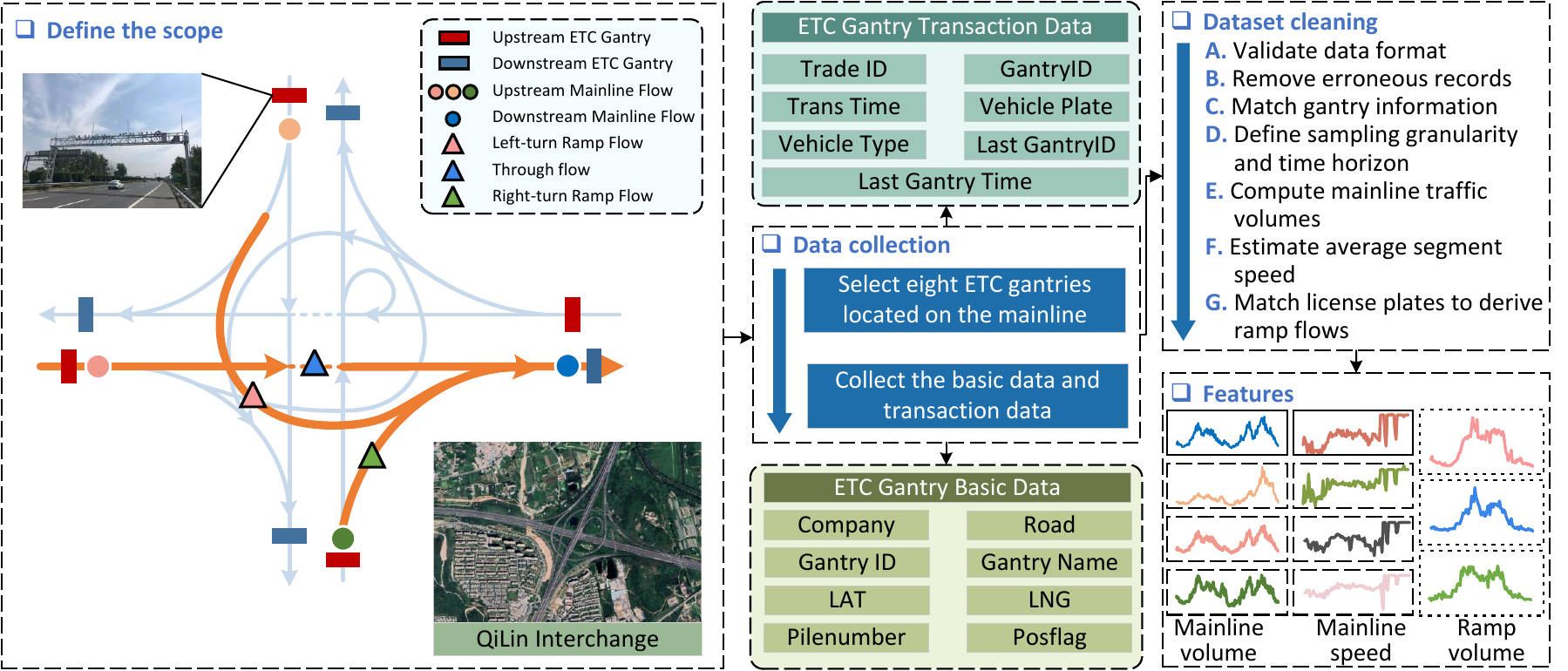}
  \caption{Data processing workflow, including the determination of the data collection scope, data collection, data cleaning, and feature extraction.}
  \label{fig:dataset}
\end{figure}

To ensure the generality and applicability of the proposed method and model, three representative and complex interchanges located in the southern region of Jiangsu Province, China, are selected as the study areas for data collection and analysis, as shown in Table~\ref{details_of_selected_interchanges}. These three interchanges differ in network structure and traffic functionality, thereby providing a comprehensive testbed for evaluating the effectiveness of the proposed algorithm. In \figurename~\ref{fig:dataset}, the QiLin interchange is taken as an example to illustrate the complete data processing workflow in detail.

\begin{table}[h]
\centering
\caption{Details of Selected Interchanges}
\label{details_of_selected_interchanges}
\small
\adjustbox{width=\textwidth,center}{
\begin{tabular}{M{2.5cm} M{2.5cm} M{4cm} M{2.5cm} J{4.5cm} M{3cm}}
\toprule
\textbf{Interchange Name} & \textbf{Location} & \textbf{Connecting Expressways} & \textbf{Interchange Type} & \textbf{Primary Function} & \textbf{Map View} \\
\midrule
QiLin & Nanjing & 
G2503 Nanjing Ring Exp. \newline
G42 Shanghai-Chengdu Exp. & 
Single-Loop Cloverleaf & 
Connects Nanjing's urban core, eastern regions, and the route to Shanghai; handles major urban-intercity traffic conversion. & 
\includegraphics[width=3cm]{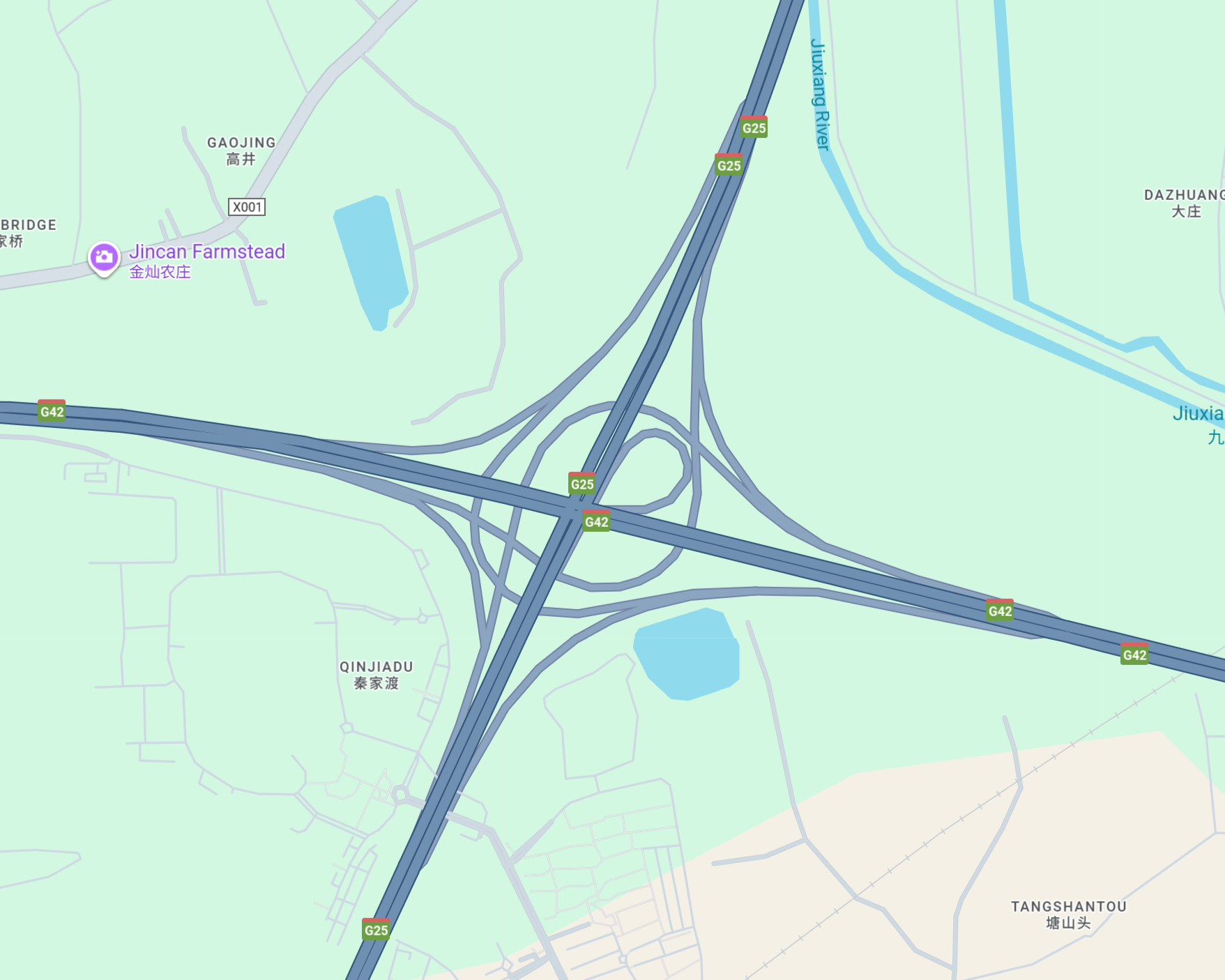} \\

DanYangXinQu & Zhenjiang & 
G42 Shanghai-Chengdu Exp. \newline
S35 Fuyang-Liyang Exp. & 
Three-Loop Cloverleaf & 
Manages traffic conversion between the east-west Shanghai-Nanjing corridor and north-south routes to central/northern Jiangsu. & 
\includegraphics[width=3cm]{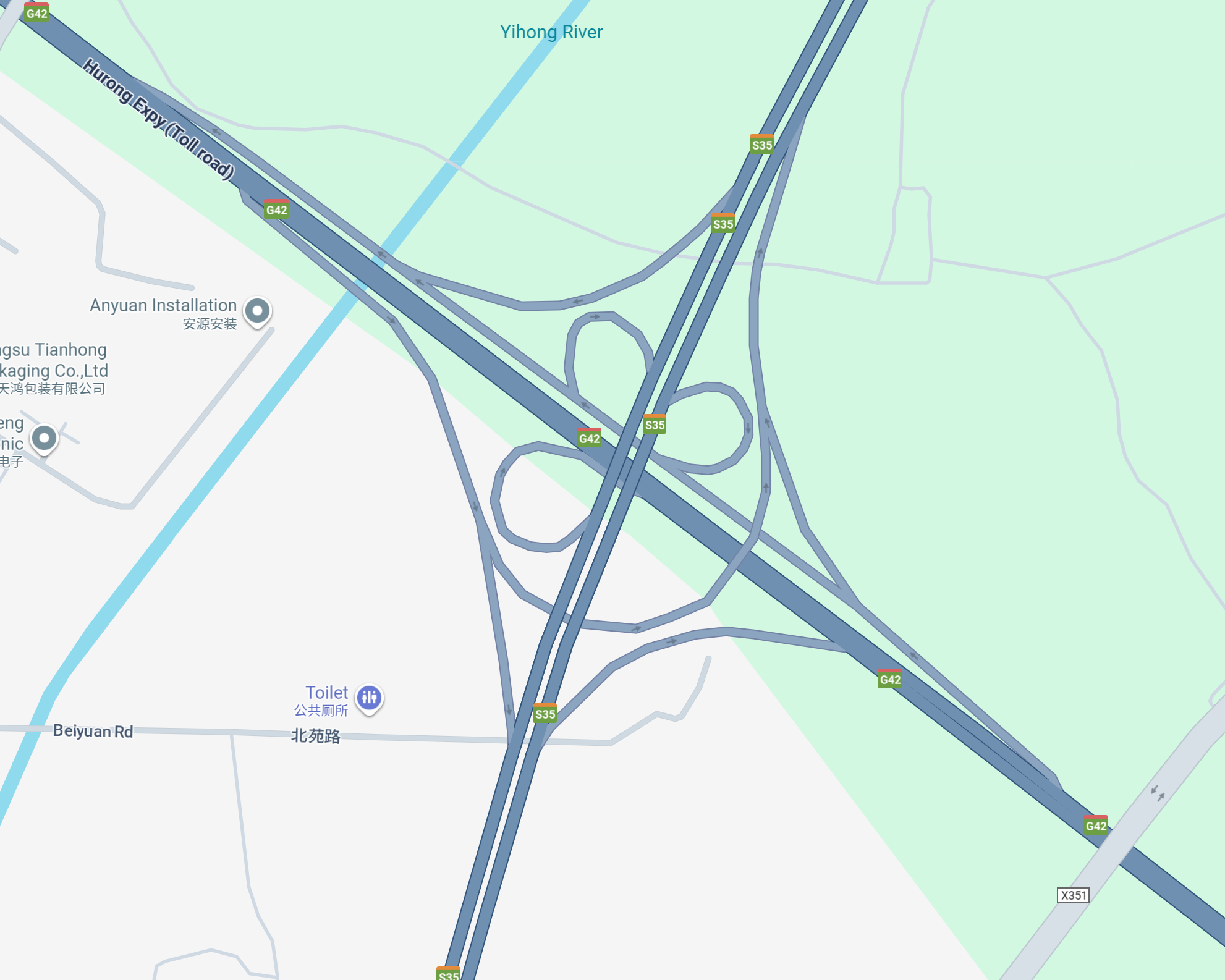} \\

XueBu & Changzhou & 
G4221 Shanghai-Wuhan Exp. \newline
G4011 Yangzhou-Liyang Exp. & 
Four-Loop Cloverleaf & 
Facilitates traffic connections between the Shanghai-Nanjing corridor and the southern Jiangsu / northern Zhejiang regions. & 
\includegraphics[width=3cm]{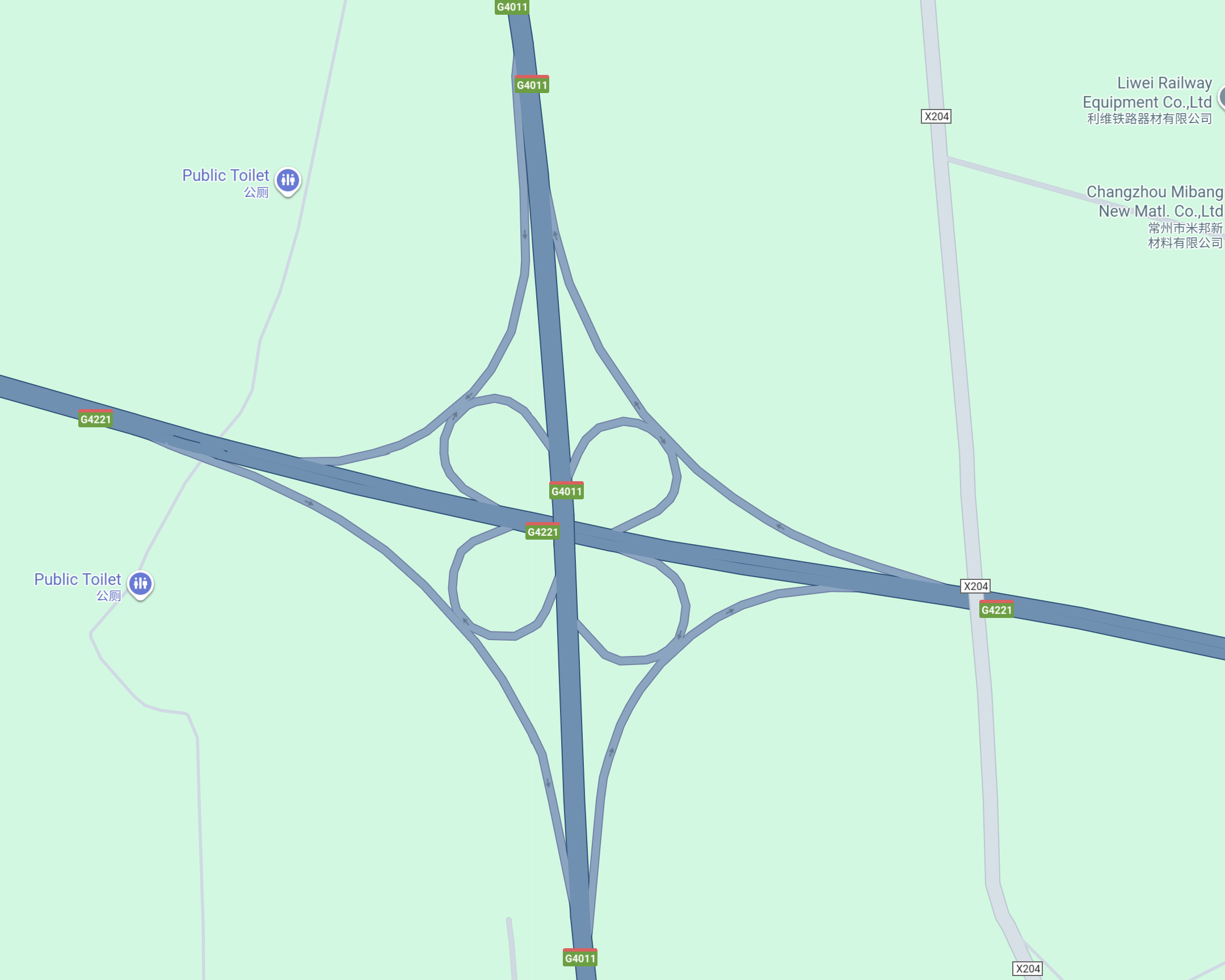} \\
\bottomrule
\end{tabular}
}
\end{table}

\subsection{Data collection}
\label{Data collection}
To comprehensively capture the traffic dynamics of the interchange, eight ETC gantries located on the upstream and downstream mainlines closest to the interchange center are selected as data collection points. Two types of data are collected: (1) basic gantry information and (2) ETC transaction records. Basic gantry information provides static attributes of each collection point, including the operating company, road segment information, gantry ID, gantry name, latitude and longitude coordinates, highway kilometer marker, and direction of travel (upstream or downstream). This information provides the geospatial foundation for subsequent calculations of segment distance and traffic record matching. ETC transaction records constitute the core dynamic data of this study. They contain the passage information of every vehicle observed at the selected gantries. Key fields include the desensitized license plate number, passage time with second-level precision, and gantry ID. These data serve as the raw basis for computing traffic volumes, speeds, and ramp flows.

\subsection{Data cleaning}
\label{Data cleaning}

The raw ETC transaction records are discrete event logs, which cannot be directly used for time series analysis. Therefore, the following data processing and feature engineering procedures are designed to transform the raw records into structured traffic features. First, the format of the acquired data is validated to ensure that all records follow a consistent structure and contain complete fields, which lays the foundation for automated processing. Erroneous records, such as timestamps outside the valid range or duplicate records of the same vehicle within the same second, are then removed. Next, each ETC record is matched with the corresponding gantry information using the gantry ID as the key index so that accurate geographical and road segment attributes can be assigned. Afterward, continuous event data are aggregated into discrete time series by defining the sampling intervals and splitting the study period into time windows. Mainline traffic volumes are computed by counting the number of vehicles passing through each gantry during every time window. To estimate mainline speed, the same license plate is matched across two adjacent gantries, the vehicle travel time over the segment is calculated, and the distance between gantries (obtained from kilometer markers) is divided by the travel time. The arithmetic mean of all vehicle speeds within a window is taken as the segment's average speed. Finally, ramp flows, which are a key feature in this study, are inferred by matching license plates observed at upstream and downstream gantries, since ramps typically lack ETC gantries. We acknowledge that this matching process is an estimation and may introduce minor measurement noise. However, given the high-quality monitoring and reliable vehicle identification of the ETC system, this resulting noise is expected to be relatively low. Through this complete process, three core time series feature sets are obtained: mainline volume, mainline speed, and ramp volume.

In this study, the data covers the period from September 7, 2024, to September 29, 2024, totaling 23 days. After data cleaning and preprocessing, a total of 3,962,597 valid vehicle passage records are obtained. Sampling intervals of 3 min, 5 min \cite{WOS:001347142804014, shao2024exploring}, and 10 min are adopted. Multiple sampling intervals are employed to assess the model’s performance under varying temporal granularities. A z-score normalization method is applied. The dataset is split into training, validation, and test sets in a ratio of 17:3:3, where the first 17 days are used as the training set, the middle 3 days as the validation set, and the last 3 days as the test set.

\section{Experiments and results}
\label{Experiments and results}
This section presents a comprehensive evaluation of the proposed STDAE model on the datasets from Section~\ref{Dataset and processing}. GWNet is selected as the downstream prediction model \cite{ijcai2019p264}, forming the combined STDAEGWNET model. Firstly, the overall performance of the STDAEGWNET model is evaluated through comparison with several state-of-the-art models. Subsequently, ablation experiments are conducted to validate the effectiveness of the pre-training approach, followed by a deep analysis of the experimental results.

\begin{table}[!htbp]
\centering
\begin{minipage}{\textwidth}
\centering
\caption{QiLin—Overall Performance across Different Sampling Intervals}
\label{tab:qilin_overall}
\renewcommand{\arraystretch}{1}
\resizebox{\textwidth}{!}{%
\begin{tabular}{l||ccc|ccc|ccc|c}
\toprule[1.5pt]
\multirow{2}{*}{\textbf{Model}} & \multicolumn{3}{c|}{\textbf{3min}} & \multicolumn{3}{c|}{\textbf{5min}} & \multicolumn{3}{c|}{\textbf{10min}} & \multirow{2}{*}{\textbf{Avg Rank $\downarrow$}} \\
\cmidrule(lr){2-4} \cmidrule(lr){5-7} \cmidrule(lr){8-10}
 & \textbf{MAE $\downarrow$} & \textbf{MAPE $\downarrow$} & \textbf{RMSE $\downarrow$} & \textbf{MAE $\downarrow$} & \textbf{MAPE $\downarrow$} & \textbf{RMSE $\downarrow$} & \textbf{MAE $\downarrow$} & \textbf{MAPE $\downarrow$} & \textbf{RMSE $\downarrow$} &  \\
\hline\hline
\textbf{GRU} & {\large 5.66} {\scriptsize $\pm$ 0.01} & {\large 0.73} {\scriptsize $\pm$ 0.01} & {\large 8.79} {\scriptsize $\pm$ 0.03} & {\large 8.81} {\scriptsize $\pm$ 0.04} & {\large 0.71} {\scriptsize $\pm$ 0.01} & {\large 13.90} {\scriptsize $\pm$ 0.07} & {\large 17.20} {\scriptsize $\pm$ 0.11} & {\large 0.69} {\scriptsize $\pm$ 0.01} & {\large 26.89} {\scriptsize $\pm$ 0.23} & 11.56 \\
\textbf{TCN} & {\large 5.63} {\scriptsize $\pm$ 0.02} & {\large 0.73} {\scriptsize $\pm$ 0.01} & {\large 8.68} {\scriptsize $\pm$ 0.03} & {\large 8.82} {\scriptsize $\pm$ 0.07} & {\large 0.72} {\scriptsize $\pm$ 0.01} & {\large 13.91} {\scriptsize $\pm$ 0.11} & {\large 17.37} {\scriptsize $\pm$ 0.24} & {\large 0.71} {\scriptsize $\pm$ 0.01} & {\large 27.03} {\scriptsize $\pm$ 0.51} & 12.11 \\
\cmidrule{1-11}
\textbf{AGCRN} & {\large 5.02} {\scriptsize $\pm$ 0.07} & \underline{{\large 0.61} {\scriptsize $\pm$ 0.01}} & {\large 7.92} {\scriptsize $\pm$ 0.13} & {\large 7.42} {\scriptsize $\pm$ 0.10} & {\large 0.55} {\scriptsize $\pm$ 0.02} & {\large 12.03} {\scriptsize $\pm$ 0.18} & {\large 14.04} {\scriptsize $\pm$ 0.25} & {\large 0.51} {\scriptsize $\pm$ 0.01} & {\large 22.73} {\scriptsize $\pm$ 0.32} & 6.33 \\
\textbf{DCRNN} & {\large 5.11} {\scriptsize $\pm$ 0.04} & {\large 0.64} {\scriptsize $\pm$ 0.01} & {\large 8.03} {\scriptsize $\pm$ 0.08} & {\large 7.55} {\scriptsize $\pm$ 0.08} & {\large 0.58} {\scriptsize $\pm$ 0.01} & {\large 12.36} {\scriptsize $\pm$ 0.24} & {\large 14.12} {\scriptsize $\pm$ 0.24} & {\large 0.55} {\scriptsize $\pm$ 0.02} & {\large 22.80} {\scriptsize $\pm$ 0.52} & 9.44 \\
\textbf{MTGNN} & {\large 6.53} {\scriptsize $\pm$ 0.61} & {\large 0.76} {\scriptsize $\pm$ 0.09} & {\large 9.88} {\scriptsize $\pm$ 0.90} & {\large 10.17} {\scriptsize $\pm$ 0.98} & {\large 0.71} {\scriptsize $\pm$ 0.06} & {\large 15.75} {\scriptsize $\pm$ 1.87} & {\large 19.12} {\scriptsize $\pm$ 1.04} & {\large 0.76} {\scriptsize $\pm$ 0.07} & {\large 28.78} {\scriptsize $\pm$ 1.57} & 13.56 \\
\textbf{STPGNN} & {\large 5.06} {\scriptsize $\pm$ 0.01} & {\large 0.67} {\scriptsize $\pm$ 0.01} & {\large 7.80} {\scriptsize $\pm$ 0.03} & {\large 7.32} {\scriptsize $\pm$ 0.05} & {\large 0.61} {\scriptsize $\pm$ 0.01} & \underline{{\large 11.59} {\scriptsize $\pm$ 0.09}} & {\large 13.31} {\scriptsize $\pm$ 0.05} & {\large 0.56} {\scriptsize $\pm$ 0.01} & {\large 21.03} {\scriptsize $\pm$ 0.12} & 5.89 \\
\textbf{TGCN} & {\large 5.89} {\scriptsize $\pm$ 0.09} & {\large 0.72} {\scriptsize $\pm$ 0.02} & {\large 9.23} {\scriptsize $\pm$ 0.18} & {\large 8.97} {\scriptsize $\pm$ 0.14} & {\large 0.68} {\scriptsize $\pm$ 0.02} & {\large 14.25} {\scriptsize $\pm$ 0.22} & {\large 19.26} {\scriptsize $\pm$ 0.26} & {\large 0.75} {\scriptsize $\pm$ 0.04} & {\large 30.76} {\scriptsize $\pm$ 0.50} & 12.78 \\
\textbf{STGCN} & {\large 5.08} {\scriptsize $\pm$ 0.02} & {\large 0.61} {\scriptsize $\pm$ 0.01} & {\large 7.93} {\scriptsize $\pm$ 0.04} & {\large 7.45} {\scriptsize $\pm$ 0.03} & {\large 0.55} {\scriptsize $\pm$ 0.01} & {\large 12.09} {\scriptsize $\pm$ 0.06} & {\large 13.63} {\scriptsize $\pm$ 0.11} & {\large 0.51} {\scriptsize $\pm$ 0.01} & {\large 22.13} {\scriptsize $\pm$ 0.23} & 6.67 \\
\textbf{D2STGNN} & \underline{{\large 4.95} {\scriptsize $\pm$ 0.04}} & {\large 0.61} {\scriptsize $\pm$ 0.01} & \underline{{\large 7.76} {\scriptsize $\pm$ 0.09}} & {\large 7.34} {\scriptsize $\pm$ 0.08} & {\large 0.54} {\scriptsize $\pm$ 0.01} & {\large 12.02} {\scriptsize $\pm$ 0.24} & {\large 12.99} {\scriptsize $\pm$ 0.12} & {\large 0.48} {\scriptsize $\pm$ 0.01} & {\large 21.06} {\scriptsize $\pm$ 0.24} & 3.56 \\
\cmidrule{1-11}
\textbf{GWNet} & {\large 4.96} {\scriptsize $\pm$ 0.03} & {\large 0.62} {\scriptsize $\pm$ 0.01} & {\large 7.78} {\scriptsize $\pm$ 0.06} & \underline{{\large 7.18} {\scriptsize $\pm$ 0.08}} & \underline{{\large 0.54} {\scriptsize $\pm$ 0.01}} & {\large 11.72} {\scriptsize $\pm$ 0.16} & \underline{{\large 12.86} {\scriptsize $\pm$ 0.21}} & \underline{{\large 0.48} {\scriptsize $\pm$ 0.01}} & \underline{{\large 20.71} {\scriptsize $\pm$ 0.40}} & 2.89 \\
\textbf{STNorm} & {\large 5.03} {\scriptsize $\pm$ 0.03} & {\large 0.62} {\scriptsize $\pm$ 0.01} & {\large 7.86} {\scriptsize $\pm$ 0.08} & {\large 7.33} {\scriptsize $\pm$ 0.07} & {\large 0.55} {\scriptsize $\pm$ 0.01} & {\large 11.79} {\scriptsize $\pm$ 0.17} & {\large 13.44} {\scriptsize $\pm$ 0.25} & {\large 0.50} {\scriptsize $\pm$ 0.02} & {\large 21.31} {\scriptsize $\pm$ 0.41} & 4.89 \\
\cmidrule{1-11}
\textbf{STAEformer} & {\large 5.10} {\scriptsize $\pm$ 0.01} & {\large 0.62} {\scriptsize $\pm$ 0.01} & {\large 7.98} {\scriptsize $\pm$ 0.04} & {\large 7.39} {\scriptsize $\pm$ 0.09} & {\large 0.55} {\scriptsize $\pm$ 0.01} & {\large 11.97} {\scriptsize $\pm$ 0.20} & {\large 13.24} {\scriptsize $\pm$ 0.22} & {\large 0.49} {\scriptsize $\pm$ 0.02} & {\large 21.59} {\scriptsize $\pm$ 0.44} & 6.11 \\
\textbf{\small iTransformer} & {\large 5.07} {\scriptsize $\pm$ 0.01} & {\large 0.62} {\scriptsize $\pm$ 0.00} & {\large 7.95} {\scriptsize $\pm$ 0.01} & {\large 7.72} {\scriptsize $\pm$ 0.06} & {\large 0.57} {\scriptsize $\pm$ 0.01} & {\large 12.31} {\scriptsize $\pm$ 0.11} & {\large 14.37} {\scriptsize $\pm$ 0.15} & {\large 0.54} {\scriptsize $\pm$ 0.01} & {\large 22.72} {\scriptsize $\pm$ 0.33} & 8.22 \\
\cmidrule{1-11}
\cellcolor{gray!25}\textbf{STDAEGWNET} & \cellcolor{gray!25}\textbf{{\large 4.89**} {\scriptsize $\pm$ 0.02}} & \cellcolor{gray!25}\textbf{{\large 0.60} {\scriptsize $\pm$ 0.01}} & \cellcolor{gray!25}\textbf{{\large 7.65**} {\scriptsize $\pm$ 0.04}} & \cellcolor{gray!25}\textbf{{\large 7.01**} {\scriptsize $\pm$ 0.04}} & \cellcolor{gray!25}\textbf{{\large 0.54} {\scriptsize $\pm$ 0.01}} & \cellcolor{gray!25}\textbf{{\large 11.29**} {\scriptsize $\pm$ 0.10}} & \cellcolor{gray!25}\textbf{{\large 12.77} {\scriptsize $\pm$ 0.13}} & \cellcolor{gray!25}\textbf{{\large 0.47} {\scriptsize $\pm$ 0.02}} & \cellcolor{gray!25}\textbf{{\large 20.58} {\scriptsize $\pm$ 0.25}} & \cellcolor{gray!25}\textbf{1.00} \\
\bottomrule[1.5pt]
\end{tabular}%
}
\end{minipage}
\vspace{0.5em}
\begin{minipage}{\textwidth}
\footnotesize
\textbf{Note:} Use bold to indicate the best-performing model and underline to indicate the second-best model. * indicates $p < 0.05$; ** indicates $p < 0.01$ for significance test between the best and second-best models. Avg Rank shows the average ranking of each model across all sampling intervals and metrics.
\end{minipage}
\vspace{0.0em} 
\begin{minipage}{\textwidth}
\centering
\caption{DanYangXinQu—Overall Performance across Different Sampling Intervals}
\label{tab:danyangxinqu_overall}
\renewcommand{\arraystretch}{1}
\resizebox{\textwidth}{!}{%
\begin{tabular}{l||ccc|ccc|ccc|c}
\toprule[1.5pt]
\multirow{2}{*}{\textbf{Model}} & \multicolumn{3}{c|}{\textbf{3min}} & \multicolumn{3}{c|}{\textbf{5min}} & \multicolumn{3}{c|}{\textbf{10min}} & \multirow{2}{*}{\textbf{Avg Rank $\downarrow$}} \\
\cmidrule(lr){2-4} \cmidrule(lr){5-7} \cmidrule(lr){8-10}
 & \textbf{MAE $\downarrow$} & \textbf{MAPE $\downarrow$} & \textbf{RMSE $\downarrow$} & \textbf{MAE $\downarrow$} & \textbf{MAPE $\downarrow$} & \textbf{RMSE $\downarrow$} & \textbf{MAE $\downarrow$} & \textbf{MAPE $\downarrow$} & \textbf{RMSE $\downarrow$} &  \\
\hline\hline
\textbf{GRU} & {\large 6.88} {\scriptsize $\pm$ 0.02} & {\large 0.63} {\scriptsize $\pm$ 0.01} & {\large 14.11} {\scriptsize $\pm$ 0.07} & {\large 10.61} {\scriptsize $\pm$ 0.16} & {\large 0.67} {\scriptsize $\pm$ 0.01} & {\large 23.00} {\scriptsize $\pm$ 0.39} & {\large 18.50} {\scriptsize $\pm$ 0.30} & {\large 0.77} {\scriptsize $\pm$ 0.01} & {\large 42.46} {\scriptsize $\pm$ 0.74} & 11.67 \\
\textbf{TCN} & {\large 6.96} {\scriptsize $\pm$ 0.04} & {\large 0.62} {\scriptsize $\pm$ 0.01} & {\large 14.26} {\scriptsize $\pm$ 0.11} & {\large 10.85} {\scriptsize $\pm$ 0.16} & {\large 0.66} {\scriptsize $\pm$ 0.01} & {\large 23.72} {\scriptsize $\pm$ 0.44} & {\large 18.91} {\scriptsize $\pm$ 0.25} & {\large 0.77} {\scriptsize $\pm$ 0.01} & {\large 43.15} {\scriptsize $\pm$ 0.57} & 12.00 \\
\cmidrule{1-11}
\textbf{AGCRN} & {\large 5.91} {\scriptsize $\pm$ 0.12} & \textbf{{\large 0.51} {\scriptsize $\pm$ 0.01}} & {\large 12.66} {\scriptsize $\pm$ 0.23} & {\large 8.90} {\scriptsize $\pm$ 0.13} & {\large 0.52} {\scriptsize $\pm$ 0.02} & {\large 20.04} {\scriptsize $\pm$ 0.38} & {\large 15.54} {\scriptsize $\pm$ 0.44} & {\large 0.56} {\scriptsize $\pm$ 0.02} & {\large 36.68} {\scriptsize $\pm$ 1.10} & 5.56 \\
\textbf{DCRNN} & {\large 6.13} {\scriptsize $\pm$ 0.09} & {\large 0.63} {\scriptsize $\pm$ 0.02} & {\large 13.09} {\scriptsize $\pm$ 0.26} & {\large 9.26} {\scriptsize $\pm$ 0.14} & {\large 0.61} {\scriptsize $\pm$ 0.02} & {\large 20.63} {\scriptsize $\pm$ 0.40} & {\large 16.40} {\scriptsize $\pm$ 0.36} & {\large 0.65} {\scriptsize $\pm$ 0.03} & {\large 38.38} {\scriptsize $\pm$ 0.95} & 9.78 \\
\textbf{MTGNN} & {\large 8.51} {\scriptsize $\pm$ 0.53} & {\large 0.80} {\scriptsize $\pm$ 0.05} & {\large 16.85} {\scriptsize $\pm$ 0.89} & {\large 12.97} {\scriptsize $\pm$ 0.99} & {\large 0.84} {\scriptsize $\pm$ 0.09} & {\large 28.40} {\scriptsize $\pm$ 2.06} & {\large 22.08} {\scriptsize $\pm$ 1.31} & {\large 0.85} {\scriptsize $\pm$ 0.05} & {\large 46.18} {\scriptsize $\pm$ 2.04} & 14.00 \\
\textbf{STPGNN} & {\large 5.79} {\scriptsize $\pm$ 0.03} & {\large 0.58} {\scriptsize $\pm$ 0.01} & {\large 12.33} {\scriptsize $\pm$ 0.09} & {\large 8.80} {\scriptsize $\pm$ 0.08} & {\large 0.58} {\scriptsize $\pm$ 0.01} & {\large 19.73} {\scriptsize $\pm$ 0.23} & {\large 14.69} {\scriptsize $\pm$ 0.12} & {\large 0.58} {\scriptsize $\pm$ 0.01} & \textbf{{\large 34.58} {\scriptsize $\pm$ 0.28}} & 4.89 \\
\textbf{TGCN} & {\large 6.93} {\scriptsize $\pm$ 0.13} & {\large 0.60} {\scriptsize $\pm$ 0.02} & {\large 14.58} {\scriptsize $\pm$ 0.29} & {\large 10.30} {\scriptsize $\pm$ 0.25} & {\large 0.66} {\scriptsize $\pm$ 0.04} & {\large 22.71} {\scriptsize $\pm$ 0.41} & {\large 19.26} {\scriptsize $\pm$ 1.00} & {\large 0.83} {\scriptsize $\pm$ 0.08} & {\large 45.61} {\scriptsize $\pm$ 2.24} & 12.00 \\
\textbf{STGCN} & {\large 5.89} {\scriptsize $\pm$ 0.04} & \underline{{\large 0.51} {\scriptsize $\pm$ 0.01}} & {\large 12.56} {\scriptsize $\pm$ 0.11} & {\large 9.04} {\scriptsize $\pm$ 0.10} & \underline{{\large 0.51} {\scriptsize $\pm$ 0.00}} & {\large 20.04} {\scriptsize $\pm$ 0.27} & {\large 15.61} {\scriptsize $\pm$ 0.31} & \underline{{\large 0.54} {\scriptsize $\pm$ 0.01}} & {\large 36.45} {\scriptsize $\pm$ 0.62} & 4.89 \\
\textbf{D2STGNN} & {\large 5.77} {\scriptsize $\pm$ 0.05} & {\large 0.53} {\scriptsize $\pm$ 0.01} & {\large 12.40} {\scriptsize $\pm$ 0.15} & \textbf{{\large 8.58} {\scriptsize $\pm$ 0.14}} & \textbf{{\large 0.50} {\scriptsize $\pm$ 0.01}} & \textbf{{\large 19.32} {\scriptsize $\pm$ 0.32}} & \textbf{{\large 14.65} {\scriptsize $\pm$ 0.31}} & {\large 0.54} {\scriptsize $\pm$ 0.03} & \underline{{\large 34.87} {\scriptsize $\pm$ 0.66}} & 2.56 \\
\cmidrule{1-11}
\textbf{GWNet} & \underline{{\large 5.66} {\scriptsize $\pm$ 0.06}} & {\large 0.52} {\scriptsize $\pm$ 0.02} & \underline{{\large 12.17} {\scriptsize $\pm$ 0.17}} & {\large 8.75} {\scriptsize $\pm$ 0.09} & {\large 0.55} {\scriptsize $\pm$ 0.02} & {\large 19.72} {\scriptsize $\pm$ 0.30} & {\large 15.01} {\scriptsize $\pm$ 0.19} & {\large 0.54} {\scriptsize $\pm$ 0.01} & {\large 35.54} {\scriptsize $\pm$ 0.46} & 3.78 \\
\textbf{STNorm} & {\large 5.93} {\scriptsize $\pm$ 0.10} & {\large 0.51} {\scriptsize $\pm$ 0.01} & {\large 12.70} {\scriptsize $\pm$ 0.25} & {\large 8.77} {\scriptsize $\pm$ 0.12} & {\large 0.53} {\scriptsize $\pm$ 0.01} & {\large 19.73} {\scriptsize $\pm$ 0.35} & {\large 15.30} {\scriptsize $\pm$ 0.17} & {\large 0.56} {\scriptsize $\pm$ 0.01} & {\large 36.00} {\scriptsize $\pm$ 0.42} & 5.33 \\
\cmidrule{1-11}
\textbf{STAEformer} & {\large 6.10} {\scriptsize $\pm$ 0.05} & {\large 0.54} {\scriptsize $\pm$ 0.01} & {\large 13.05} {\scriptsize $\pm$ 0.15} & {\large 9.05} {\scriptsize $\pm$ 0.10} & {\large 0.53} {\scriptsize $\pm$ 0.02} & {\large 20.31} {\scriptsize $\pm$ 0.38} & {\large 15.18} {\scriptsize $\pm$ 0.54} & {\large 0.60} {\scriptsize $\pm$ 0.04} & {\large 35.40} {\scriptsize $\pm$ 1.18} & 6.89 \\
\textbf{\small iTransformer} & {\large 6.14} {\scriptsize $\pm$ 0.03} & {\large 0.58} {\scriptsize $\pm$ 0.01} & {\large 13.24} {\scriptsize $\pm$ 0.11} & {\large 9.29} {\scriptsize $\pm$ 0.10} & {\large 0.60} {\scriptsize $\pm$ 0.01} & {\large 21.28} {\scriptsize $\pm$ 0.30} & {\large 16.03} {\scriptsize $\pm$ 0.22} & {\large 0.68} {\scriptsize $\pm$ 0.02} & {\large 38.33} {\scriptsize $\pm$ 0.46} & 9.56 \\
\cmidrule{1-11}
\cellcolor{gray!25}\textbf{STDAEGWNET} & \cellcolor{gray!25}\textbf{{\large 5.61*} {\scriptsize $\pm$ 0.02}} & \cellcolor{gray!25}{\large 0.51} {\scriptsize $\pm$ 0.01} & \cellcolor{gray!25}\textbf{{\large 12.07} {\scriptsize $\pm$ 0.11}} & \cellcolor{gray!25}\underline{{\large 8.59} {\scriptsize $\pm$ 0.08}} & \cellcolor{gray!25}{\large 0.53} {\scriptsize $\pm$ 0.02} & \cellcolor{gray!25}\underline{{\large 19.57} {\scriptsize $\pm$ 0.25}} & \cellcolor{gray!25}\underline{{\large 14.69} {\scriptsize $\pm$ 0.07}} & \cellcolor{gray!25}\textbf{{\large 0.52*} {\scriptsize $\pm$ 0.01}} & \cellcolor{gray!25}{\large 35.37} {\scriptsize $\pm$ 0.32} & \cellcolor{gray!25}\textbf{2.11} \\
\bottomrule[1.5pt]
\end{tabular}%
}
\end{minipage}
\vspace{0.5em}
\begin{minipage}{\textwidth}
\footnotesize
\textbf{Note:} Use bold to indicate the best-performing model and underline to indicate the second-best model. * indicates $p < 0.05$; ** indicates $p < 0.01$ for significance test between the best and second-best models. Avg Rank shows the average ranking of each model across all sampling intervals and metrics.
\end{minipage}
\vspace{0.0em}
\begin{minipage}{\textwidth}
\centering
\caption{XueBu—Overall Performance across Different Sampling Intervals}
\label{tab:xuebu_overall}
\renewcommand{\arraystretch}{1}
\resizebox{\textwidth}{!}{%
\begin{tabular}{l||ccc|ccc|ccc|c}
\toprule[1.5pt]
\multirow{2}{*}{\textbf{Model}} & \multicolumn{3}{c|}{\textbf{3min}} & \multicolumn{3}{c|}{\textbf{5min}} & \multicolumn{3}{c|}{\textbf{10min}} & \multirow{2}{*}{\textbf{Avg Rank $\downarrow$}} \\
\cmidrule(lr){2-4} \cmidrule(lr){5-7} \cmidrule(lr){8-10}
 & \textbf{MAE $\downarrow$} & \textbf{MAPE $\downarrow$} & \textbf{RMSE $\downarrow$} & \textbf{MAE $\downarrow$} & \textbf{MAPE $\downarrow$} & \textbf{RMSE $\downarrow$} & \textbf{MAE $\downarrow$} & \textbf{MAPE $\downarrow$} & \textbf{RMSE $\downarrow$} &  \\
\hline\hline
\textbf{GRU} & {\large 5.42} {\scriptsize $\pm$ 0.03} & {\large 0.64} {\scriptsize $\pm$ 0.01} & {\large 9.57} {\scriptsize $\pm$ 0.09} & {\large 8.30} {\scriptsize $\pm$ 0.08} & {\large 0.67} {\scriptsize $\pm$ 0.01} & {\large 15.13} {\scriptsize $\pm$ 0.18} & {\large 15.66} {\scriptsize $\pm$ 0.18} & {\large 0.87} {\scriptsize $\pm$ 0.02} & {\large 29.92} {\scriptsize $\pm$ 0.48} & 12.78 \\
\textbf{TCN} & {\large 5.37} {\scriptsize $\pm$ 0.02} & {\large 0.64} {\scriptsize $\pm$ 0.01} & {\large 9.44} {\scriptsize $\pm$ 0.05} & {\large 8.29} {\scriptsize $\pm$ 0.07} & {\large 0.67} {\scriptsize $\pm$ 0.01} & {\large 14.98} {\scriptsize $\pm$ 0.13} & {\large 15.15} {\scriptsize $\pm$ 0.11} & {\large 0.87} {\scriptsize $\pm$ 0.02} & {\large 28.63} {\scriptsize $\pm$ 0.27} & 11.78 \\
\cmidrule{1-11}
\textbf{AGCRN} & {\large 4.79} {\scriptsize $\pm$ 0.07} & {\large 0.58} {\scriptsize $\pm$ 0.01} & {\large 8.44} {\scriptsize $\pm$ 0.11} & {\large 7.37} {\scriptsize $\pm$ 0.11} & {\large 0.60} {\scriptsize $\pm$ 0.02} & {\large 13.42} {\scriptsize $\pm$ 0.14} & {\large 13.85} {\scriptsize $\pm$ 0.51} & {\large 0.74} {\scriptsize $\pm$ 0.04} & {\large 26.62} {\scriptsize $\pm$ 1.07} & 8.67 \\
\textbf{DCRNN} & {\large 4.86} {\scriptsize $\pm$ 0.07} & {\large 0.66} {\scriptsize $\pm$ 0.02} & {\large 8.38} {\scriptsize $\pm$ 0.11} & {\large 7.50} {\scriptsize $\pm$ 0.19} & {\large 0.64} {\scriptsize $\pm$ 0.04} & {\large 13.63} {\scriptsize $\pm$ 0.53} & {\large 13.73} {\scriptsize $\pm$ 0.34} & {\large 0.76} {\scriptsize $\pm$ 0.04} & {\large 25.47} {\scriptsize $\pm$ 0.74} & 9.33 \\
\textbf{MTGNN} & {\large 6.04} {\scriptsize $\pm$ 0.36} & {\large 0.67} {\scriptsize $\pm$ 0.05} & {\large 10.30} {\scriptsize $\pm$ 0.90} & {\large 8.70} {\scriptsize $\pm$ 0.29} & {\large 0.66} {\scriptsize $\pm$ 0.04} & {\large 14.74} {\scriptsize $\pm$ 0.52} & {\large 18.00} {\scriptsize $\pm$ 0.94} & {\large 0.94} {\scriptsize $\pm$ 0.11} & {\large 31.48} {\scriptsize $\pm$ 2.96} & 13.56 \\
\textbf{STPGNN} & {\large 4.74} {\scriptsize $\pm$ 0.03} & {\large 0.60} {\scriptsize $\pm$ 0.01} & {\large 8.26} {\scriptsize $\pm$ 0.06} & {\large 7.17} {\scriptsize $\pm$ 0.05} & {\large 0.59} {\scriptsize $\pm$ 0.01} & \underline{{\large 12.93} {\scriptsize $\pm$ 0.10}} & {\large 12.86} {\scriptsize $\pm$ 0.13} & {\large 0.67} {\scriptsize $\pm$ 0.02} & {\large 24.21} {\scriptsize $\pm$ 0.23} & 5.89 \\
\textbf{TGCN} & {\large 5.38} {\scriptsize $\pm$ 0.11} & {\large 0.63} {\scriptsize $\pm$ 0.02} & {\large 9.65} {\scriptsize $\pm$ 0.24} & {\large 7.89} {\scriptsize $\pm$ 0.24} & {\large 0.65} {\scriptsize $\pm$ 0.04} & {\large 14.45} {\scriptsize $\pm$ 0.57} & {\large 14.99} {\scriptsize $\pm$ 0.45} & {\large 0.87} {\scriptsize $\pm$ 0.06} & {\large 28.87} {\scriptsize $\pm$ 0.94} & 11.56 \\
\textbf{STGCN} & {\large 4.73} {\scriptsize $\pm$ 0.02} & {\large 0.56} {\scriptsize $\pm$ 0.01} & {\large 8.33} {\scriptsize $\pm$ 0.04} & {\large 7.23} {\scriptsize $\pm$ 0.07} & {\large 0.55} {\scriptsize $\pm$ 0.01} & {\large 13.25} {\scriptsize $\pm$ 0.11} & {\large 13.13} {\scriptsize $\pm$ 0.20} & {\large 0.62} {\scriptsize $\pm$ 0.03} & {\large 25.11} {\scriptsize $\pm$ 0.48} & 6.22 \\
\textbf{D2STGNN} & \underline{{\large 4.62} {\scriptsize $\pm$ 0.03}} & \underline{{\large 0.55} {\scriptsize $\pm$ 0.01}} & \underline{{\large 8.17} {\scriptsize $\pm$ 0.05}} & \underline{{\large 6.98} {\scriptsize $\pm$ 0.07}} & \textbf{{\large 0.50} {\scriptsize $\pm$ 0.02}} & {\large 12.95} {\scriptsize $\pm$ 0.27} & {\large 12.38} {\scriptsize $\pm$ 0.19} & \textbf{{\large 0.54} {\scriptsize $\pm$ 0.04}} & {\large 24.15} {\scriptsize $\pm$ 0.56} & 2.11 \\
\cmidrule{1-11}
\textbf{GWNet} & {\large 4.64} {\scriptsize $\pm$ 0.05} & {\large 0.55} {\scriptsize $\pm$ 0.02} & {\large 8.23} {\scriptsize $\pm$ 0.12} & {\large 7.35} {\scriptsize $\pm$ 0.14} & {\large 0.54} {\scriptsize $\pm$ 0.02} & {\large 13.99} {\scriptsize $\pm$ 0.47} & \underline{{\large 12.34} {\scriptsize $\pm$ 0.30}} & {\large 0.55} {\scriptsize $\pm$ 0.02} & {\large 24.18} {\scriptsize $\pm$ 0.88} & 4.44 \\
\textbf{STNorm} & {\large 4.69} {\scriptsize $\pm$ 0.03} & {\large 0.56} {\scriptsize $\pm$ 0.01} & {\large 8.31} {\scriptsize $\pm$ 0.08} & {\large 7.18} {\scriptsize $\pm$ 0.20} & {\large 0.53} {\scriptsize $\pm$ 0.01} & {\large 13.36} {\scriptsize $\pm$ 0.53} & {\large 12.73} {\scriptsize $\pm$ 0.25} & \underline{{\large 0.55} {\scriptsize $\pm$ 0.02}} & {\large 25.09} {\scriptsize $\pm$ 0.82} & 4.67 \\
\cmidrule{1-11}
\textbf{STAEformer} & {\large 4.71} {\scriptsize $\pm$ 0.04} & {\large 0.56} {\scriptsize $\pm$ 0.01} & {\large 8.30} {\scriptsize $\pm$ 0.07} & {\large 7.12} {\scriptsize $\pm$ 0.09} & {\large 0.53} {\scriptsize $\pm$ 0.02} & {\large 13.01} {\scriptsize $\pm$ 0.14} & {\large 12.54} {\scriptsize $\pm$ 0.14} & {\large 0.60} {\scriptsize $\pm$ 0.03} & \underline{{\large 23.97} {\scriptsize $\pm$ 0.29}} & 4.22 \\
\textbf{\small iTransformer} & {\large 5.10} {\scriptsize $\pm$ 0.02} & {\large 0.56} {\scriptsize $\pm$ 0.01} & {\large 9.10} {\scriptsize $\pm$ 0.05} & {\large 7.64} {\scriptsize $\pm$ 0.05} & {\large 0.54} {\scriptsize $\pm$ 0.01} & {\large 14.13} {\scriptsize $\pm$ 0.11} & {\large 13.68} {\scriptsize $\pm$ 0.15} & {\large 0.63} {\scriptsize $\pm$ 0.02} & {\large 26.35} {\scriptsize $\pm$ 0.26} & 8.33 \\
\cmidrule{1-11}
\cellcolor{gray!25}\textbf{STDAEGWNET} & \cellcolor{gray!25}\textbf{{\large 4.58**} {\scriptsize $\pm$ 0.01}} & \cellcolor{gray!25}\textbf{{\large 0.54*} {\scriptsize $\pm$ 0.01}} & \cellcolor{gray!25}\textbf{{\large 8.11*} {\scriptsize $\pm$ 0.07}} & \cellcolor{gray!25}\textbf{{\large 6.89**} {\scriptsize $\pm$ 0.04}} & \cellcolor{gray!25}\underline{{\large 0.52} {\scriptsize $\pm$ 0.01}} & \cellcolor{gray!25}\textbf{{\large 12.89} {\scriptsize $\pm$ 0.14}} & \cellcolor{gray!25}\textbf{{\large 12.30} {\scriptsize $\pm$ 0.05}} & \cellcolor{gray!25}{\large 0.56} {\scriptsize $\pm$ 0.02} & \cellcolor{gray!25}\textbf{{\large 23.70*} {\scriptsize $\pm$ 0.07}} & \cellcolor{gray!25}\textbf{1.44} \\
\bottomrule[1.5pt]
\end{tabular}%
}
\end{minipage}
\vspace{0.5em}
\begin{minipage}{\textwidth}
\footnotesize
\textbf{Note:} Use bold to indicate the best-performing model and underline to indicate the second-best model. * indicates $p < 0.05$; ** indicates $p < 0.01$ for significance test between the best and second-best models. Avg Rank shows the average ranking of each model across all sampling intervals and metrics.
\end{minipage}
\end{table}

\subsection{Experimental setup}
\label{Experimental setup}

\textbf{Baselines.} In this section, STDAEGWNET is compared with four groups of representative benchmark models. The \textit{time-series models} include GRU and TCN. The \textit{spatio-temporal GNN models} comprise AGCRN \cite{bai2020adaptive}, DCRNN \cite{li2017diffusion}, MTGNN \cite{wu2020connecting}, STPGNN \cite{kong2024spatio}, TGCN \cite{zhao2019t}, STGCN \cite{WOS:000764175403107}, and D2STGNN \cite{shao2022decoupled}. The \textit{spatio-temporal enhanced CNN models} include GWNet \cite{ijcai2019p264} and STNorm \cite{deng2021st}. The \textit{spatio-temporal Transformer models} consist of STAEformer \cite{liu2023spatio} and iTransformer \cite{liu2024itransformer}. These baselines span traditional statistical and machine learning methods to the latest graph neural networks and Transformer-based architectures, providing a comprehensive evaluation of the spatio-temporal modeling and forecasting capabilities of STDAEGWNET. The input to these baseline models is consistent with STDAEGWNET, consisting of mainline feature time series. In addition, the performance of HimNet \cite{dong2024heterogeneity}, GWNet \cite{ijcai2019p264}, and STDMAE \cite{WOS:001347142804014} with ramp flow time series as input is compared to that of our model using mainline feature time series. The purpose is to verify whether using mainline features as input can achieve better prediction performance than directly using ramp flow sequences, thereby demonstrating the feasibility of predicting using only mainline data.

\textbf{Settings.} To determine the optimal model configuration, we conduct manual hyperparameter tuning evaluated at the 5 min sampling interval across the three datasets. We explore the long temporal horizon $T_{\text{long}} \in \{1, 2, 3\}$ days, the embedding dimension $G \in \{16, 32, 64, 96\}$, and the initial learning rate $\text{LR} \in \{0.0002, 0.002, 0.02\}$. The parameter combination that yielded the lowest Mean Absolute Error (MAE) is selected for final evaluation, as shown in \figurename~\ref{fig:HyperparameterSearch}. Consequently, in the pre-training stage, the long temporal horizon is set to $T_{\text{long}} = 1day$. In the forecasting stage, the input sequence length is fixed at $T = 12$ \cite{WOS:001347142804014,song2020spatial} and the embedding dimension is set to $G = 96$. The encoder comprises four Transformer layers and the decoder one Transformer layer, both employing four-head multi-head attention. To match the forecasting input, the patch size is $L = 12$, and $T' = 1$, retaining only the last patch from the spatial representation \( \mathbf{H}^{(S)} \) and temporal representation \( \mathbf{H}^{(T)} \). Training uses the Adam optimizer with the selected initial learning rate of 0.002 and MAE loss. Performance is evaluated by MAE, Mean Absolute Percentage Error (MAPE), and Root Mean Squared Error (RMSE). The results are averaged over 10 independent runs. All experiments are conducted on a Windows workstation equipped with a single NVIDIA V100 Tensor Core GPU. The experiments are conducted using the platform \cite{shao2024exploring}.

\subsection{Overall performance}
\label{Overall performance}
The performance of models is listed in Table~\ref{tab:qilin_overall}, Table~\ref{tab:danyangxinqu_overall}, and Table~\ref{tab:xuebu_overall}. The proposed STDAEGWNET model demonstrates significant superiority across all three interchange datasets. Moreover, for the 3 min, 5 min, and 10 min forecasting tasks, the model achieves the best or second-best performance on the MAE, MAPE, and RMSE metrics, consistently maintaining the top overall average ranking.

Specifically, the model performance is reflected in the following four aspects. \textbf{Overall Performance}. The average ranking results further validate the comprehensive superiority of STDAEGWNET. The model average ranks on the QiLin, DanYangXinQu, and XueBu datasets are 1.00, 2.11, and 1.44, respectively, all ranking first. This performance clearly surpasses existing mainstream models, including graph neural network models, whose best ranks on the three datasets reach only 2.11, and Transformer-based models, whose best rank is only 4.22. This demonstrates that the proposed pre-training and downstream prediction framework can more effectively capture the spatiotemporal dependencies between mainline and ramp traffic, thereby exhibiting outstanding performance in practical prediction tasks. \textbf{Accuracy}. In most prediction scenarios, STDAEGWNET achieves the lowest MAE, indicating that its predictions deviate least from the true values and outperform other baseline models in accuracy. At a 3 min sampling interval, the model mean absolute errors on the QiLin, DanYangXinQu, and XueBu datasets are 4.89, 5.61, and 4.58, respectively, all the lowest among the compared models and statistically significant. Even in scenarios where it does not achieve absolute optimality, performance remains close to the best, maintaining stable competitiveness. \textbf{Stability and Robustness}. The RMSE results show that STDAEGWNET generally outperforms other models, indicating concentrated prediction distributions and fewer extreme errors. On the QiLin dataset for 5 min predictions, STDAEGWNET achieves a RMSE of 11.29, significantly lower than GRU at 13.90, TCN at 13.91, and DCRNN at 12.36. On the XueBu dataset for 10 min predictions, root mean square error is 23.70, lower than D2STGNN at 24.15 and GWNet at 24.18, demonstrating the model's stability and robustness. \textbf{Generalizability}. Across three different hub datasets, STDAEGWNET consistently ranks among the top, indicating strong cross-scenario generalization and adaptability to diverse traffic characteristics. Compared with models that excel only on a single dataset, STDAEGWNET shows more robust overall performance. D2STGNN performs strongly on the XueBu and DanYangXinQu datasets with average ranks of 2.11 and 2.56 but performs relatively weaker on the QiLin dataset at 3.56. In contrast, STDAEGWNET maintains the best average ranks across all three datasets at 1.00, 2.11, and 1.44, highlighting superior robustness and consistency across scenarios.

\begin{table}[t]
\centering
\caption{Performance Comparison with Advanced Ramp Flow Prediction Models Using Ramp Flow Sequences}
\label{tab:comparison_with_ramptoramp}
{\fontfamily{ptm}\selectfont
\renewcommand{\arraystretch}{1.2}
\resizebox{\textwidth}{!}{%
\begin{tabular}{p{1.5cm}p{2.5cm}|p{1.8cm}p{1.8cm}p{1.8cm}|p{1.8cm}p{1.8cm}p{1.8cm}|p{1.8cm}p{1.8cm}p{1.8cm}}
\toprule[1.5pt]
\multirow{2}{*}{\textbf{Dataset}} & \multirow{2}{*}{\textbf{Model}} & \multicolumn{3}{c|}{\textbf{3min}} & \multicolumn{3}{c|}{\textbf{5min}} & \multicolumn{3}{c}{\textbf{10min}} \\
\cmidrule(lr){3-5} \cmidrule(lr){6-8} \cmidrule(lr){9-11}
 &  & \textbf{MAE} & \textbf{MAPE} & \textbf{RMSE} & \textbf{MAE} & \textbf{MAPE} & \textbf{RMSE} & \textbf{MAE} & \textbf{MAPE} & \textbf{RMSE} \\
\hline\hline
\multirow{4}{*}{\textbf{QiLin}} & \textbf{HimNet} & \mbox{5.10 $\pm$ {\scriptsize 0.04}} & \mbox{0.63 $\pm$ {\scriptsize 0.01}} & \mbox{8.18 $\pm$ {\scriptsize 0.09}} & \mbox{7.74 $\pm$ {\scriptsize 0.08}} & \mbox{0.56 $\pm$ {\scriptsize 0.01}} & \mbox{13.08 $\pm$ {\scriptsize 0.19}} & \mbox{13.82 $\pm$ {\scriptsize 0.33}} & \mbox{0.45 $\pm$ {\scriptsize 0.01}} & \mbox{23.46 $\pm$ {\scriptsize 0.79}} \\
 & \textbf{GWNet} & \mbox{4.78 $\pm$ {\scriptsize 0.04}} & \mbox{0.59 $\pm$ {\scriptsize 0.01}} & \mbox{7.70 $\pm$ {\scriptsize 0.09}} & \mbox{7.20 $\pm$ {\scriptsize 0.07}} & \mbox{0.53 $\pm$ {\scriptsize 0.01}} & \mbox{11.87 $\pm$ {\scriptsize 0.15}} & \mbox{12.65 $\pm$ {\scriptsize 0.20}} & \mbox{0.45 $\pm$ {\scriptsize 0.01}} & \mbox{20.53 $\pm$ {\scriptsize 0.31}} \\
 & \textbf{STDMAE} & \mbox{4.66 $\pm$ {\scriptsize 0.01}} & \mbox{0.58 $\pm$ {\scriptsize 0.00}} & \mbox{7.36 $\pm$ {\scriptsize 0.04}} & \mbox{7.05 $\pm$ {\scriptsize 0.03}} & \mbox{0.53 $\pm$ {\scriptsize 0.01}} & \mbox{11.51 $\pm$ {\scriptsize 0.07}} & \mbox{13.23 $\pm$ {\scriptsize 0.17}} & \mbox{0.50 $\pm$ {\scriptsize 0.01}} & \mbox{21.28 $\pm$ {\scriptsize 0.32}} \\
\cmidrule(lr){2-11}
 & \cellcolor{gray!25}\textbf{STDAEGWNET} & \cellcolor{gray!25}\mbox{4.89$^{1}$ $\pm$ {\scriptsize 0.02}} & \cellcolor{gray!25}\mbox{0.60$^{1}$ $\pm$ {\scriptsize 0.01}} & \cellcolor{gray!25}\mbox{7.65$^{1}$ $\pm$ {\scriptsize 0.04}} & \cellcolor{gray!25}\mbox{7.01$^{a}$ $\pm$ {\scriptsize 0.04}} & \cellcolor{gray!25}\mbox{0.54$^{1}$ $\pm$ {\scriptsize 0.01}} & \cellcolor{gray!25}\mbox{11.29$^{a}$ $\pm$ {\scriptsize 0.10}} & \cellcolor{gray!25}\mbox{12.77$^{1,3}$ $\pm$ {\scriptsize 0.13}} & \cellcolor{gray!25}\mbox{0.47$^{3}$ $\pm$ {\scriptsize 0.02}} & \cellcolor{gray!25}\mbox{20.58$^{1,3}$ $\pm$ {\scriptsize 0.25}} \\
\hline\hline
\multirow{4}{*}{\parbox{1.4cm}{\centering \textbf{DanYang}\\\textbf{XinQu}}} & \textbf{HimNet} & \mbox{6.10 $\pm$ {\scriptsize 0.10}} & \mbox{0.59 $\pm$ {\scriptsize 0.02}} & \mbox{12.97 $\pm$ {\scriptsize 0.29}} & \mbox{9.35 $\pm$ {\scriptsize 0.18}} & \mbox{0.56 $\pm$ {\scriptsize 0.03}} & \mbox{21.20 $\pm$ {\scriptsize 0.55}} & \mbox{15.72 $\pm$ {\scriptsize 0.30}} & \mbox{0.57 $\pm$ {\scriptsize 0.04}} & \mbox{37.85 $\pm$ {\scriptsize 0.82}} \\
 & \textbf{GWNet} & \mbox{5.75 $\pm$ {\scriptsize 0.07}} & \mbox{0.49 $\pm$ {\scriptsize 0.01}} & \mbox{12.35 $\pm$ {\scriptsize 0.19}} & \mbox{8.52 $\pm$ {\scriptsize 0.12}} & \mbox{0.49 $\pm$ {\scriptsize 0.01}} & \mbox{19.36 $\pm$ {\scriptsize 0.38}} & \mbox{14.51 $\pm$ {\scriptsize 0.38}} & \mbox{0.50 $\pm$ {\scriptsize 0.02}} & \mbox{35.40 $\pm$ {\scriptsize 1.15}} \\
 & \textbf{STDMAE} & \mbox{5.64 $\pm$ {\scriptsize 0.02}} & \mbox{0.48 $\pm$ {\scriptsize 0.00}} & \mbox{12.17 $\pm$ {\scriptsize 0.05}} & \mbox{8.52 $\pm$ {\scriptsize 0.05}} & \mbox{0.50 $\pm$ {\scriptsize 0.01}} & \mbox{19.30 $\pm$ {\scriptsize 0.18}} & \mbox{14.25 $\pm$ {\scriptsize 0.13}} & \mbox{0.50 $\pm$ {\scriptsize 0.01}} & \mbox{33.92 $\pm$ {\scriptsize 0.35}} \\
\cmidrule(lr){2-11}
 & \cellcolor{gray!25}\textbf{STDAEGWNET} & \cellcolor{gray!25}\mbox{5.61$^{a}$ $\pm$ {\scriptsize 0.02}} & \cellcolor{gray!25}\mbox{0.51$^{1}$ $\pm$ {\scriptsize 0.01}} & \cellcolor{gray!25}\mbox{12.07$^{a}$ $\pm$ {\scriptsize 0.11}} & \cellcolor{gray!25}\mbox{8.59$^{1}$ $\pm$ {\scriptsize 0.08}} & \cellcolor{gray!25}\mbox{0.53$^{1}$ $\pm$ {\scriptsize 0.02}} & \cellcolor{gray!25}\mbox{19.57$^{1}$ $\pm$ {\scriptsize 0.25}} & \cellcolor{gray!25}\mbox{14.69$^{1}$ $\pm$ {\scriptsize 0.07}} & \cellcolor{gray!25}\mbox{0.52$^{1}$ $\pm$ {\scriptsize 0.01}} & \cellcolor{gray!25}\mbox{35.37$^{1}$ $\pm$ {\scriptsize 0.32}} \\
\hline\hline
\multirow{4}{*}{\textbf{XueBu}} & \textbf{HimNet} & \mbox{4.94 $\pm$ {\scriptsize 0.01}} & \mbox{0.57 $\pm$ {\scriptsize 0.02}} & \mbox{8.76 $\pm$ {\scriptsize 0.03}} & \mbox{7.26 $\pm$ {\scriptsize 0.05}} & \mbox{0.52 $\pm$ {\scriptsize 0.01}} & \mbox{13.34 $\pm$ {\scriptsize 0.10}} & \mbox{12.91 $\pm$ {\scriptsize 0.16}} & \mbox{0.55 $\pm$ {\scriptsize 0.03}} & \mbox{24.81 $\pm$ {\scriptsize 0.26}} \\
 & \textbf{GWNet} & \mbox{4.75 $\pm$ {\scriptsize 0.02}} & \mbox{0.54 $\pm$ {\scriptsize 0.01}} & \mbox{8.47 $\pm$ {\scriptsize 0.05}} & \mbox{7.27 $\pm$ {\scriptsize 0.08}} & \mbox{0.55 $\pm$ {\scriptsize 0.01}} & \mbox{13.43 $\pm$ {\scriptsize 0.20}} & \mbox{12.28 $\pm$ {\scriptsize 0.15}} & \mbox{0.53 $\pm$ {\scriptsize 0.02}} & \mbox{23.95 $\pm$ {\scriptsize 0.46}} \\
 & \textbf{STDMAE} & \mbox{4.68 $\pm$ {\scriptsize 0.01}} & \mbox{0.51 $\pm$ {\scriptsize 0.00}} & \mbox{8.38 $\pm$ {\scriptsize 0.02}} & \mbox{7.00 $\pm$ {\scriptsize 0.01}} & \mbox{0.49 $\pm$ {\scriptsize 0.00}} & \mbox{12.98 $\pm$ {\scriptsize 0.05}} & \mbox{12.28 $\pm$ {\scriptsize 0.07}} & \mbox{0.54 $\pm$ {\scriptsize 0.01}} & \mbox{23.50 $\pm$ {\scriptsize 0.17}} \\
\cmidrule(lr){2-11}
 & \cellcolor{gray!25}\textbf{STDAEGWNET} & \cellcolor{gray!25}\mbox{4.58$^{a}$ $\pm$ {\scriptsize 0.01}} & \cellcolor{gray!25}\mbox{0.54$^{1}$ $\pm$ {\scriptsize 0.01}} & \cellcolor{gray!25}\mbox{8.11$^{a}$ $\pm$ {\scriptsize 0.07}} & \cellcolor{gray!25}\mbox{6.89$^{a}$ $\pm$ {\scriptsize 0.04}} & \cellcolor{gray!25}\mbox{0.52$^{2}$ $\pm$ {\scriptsize 0.01}} & \cellcolor{gray!25}\mbox{12.89$^{1,2}$ $\pm$ {\scriptsize 0.14}} & \cellcolor{gray!25}\mbox{12.30$^{1}$ $\pm$ {\scriptsize 0.05}} & \cellcolor{gray!25}\mbox{0.56 $\pm$ {\scriptsize 0.02}} & \cellcolor{gray!25}\mbox{23.70$^{1}$ $\pm$ {\scriptsize 0.07}} \\
\bottomrule[1.5pt]
\end{tabular}%
}
}
\begin{minipage}{\textwidth}
\small
\textbf{Note:} Superscripts indicate significant improvements ($p < 0.05$) of STDAEGWNET over: 
$^1$HimNet, $^2$GWNet, $^3$STDMAE, and $^{a}$all three models.
\end{minipage}
\end{table}

Table~\ref{tab:comparison_with_ramptoramp} presents a comprehensive performance comparison between the proposed STDAEGWNET model, which uses mainline data, and three advanced benchmark models, HimNet, GWNet, and STDMAE, all of which rely on historical ramp data as input. The experimental results show that our proposed STDAEGWNET model, even without using the target ramp historical flow, achieves highly competitive performance and significantly outperforms models that directly use ramp historical data in many critical scenarios.

The results clearly demonstrate that STDAEGWNET consistently and significantly outperforms HimNet. Across nearly all evaluation metrics and scenarios, the improvements over HimNet are statistically significant ($p < 0.05$), providing initial confirmation of the effectiveness of our model architecture. Furthermore, GWNet and STDMAE are two very strong benchmark models, making the comparison with them particularly valuable. The experiments reveal that in certain scenarios, models that directly leverage their own historical ramp data outperform STDAEGWNET. Nevertheless, STDAEGWNET still achieves superior performance in many cases. For example, in the 3 min forecasting tasks on the DanYangXinQu and XueBu datasets, its MAE and RMSE are significantly better than those of GWNet and STDMAE.

\begin{figure}[t]
  \centering
  \includegraphics[width=\textwidth]{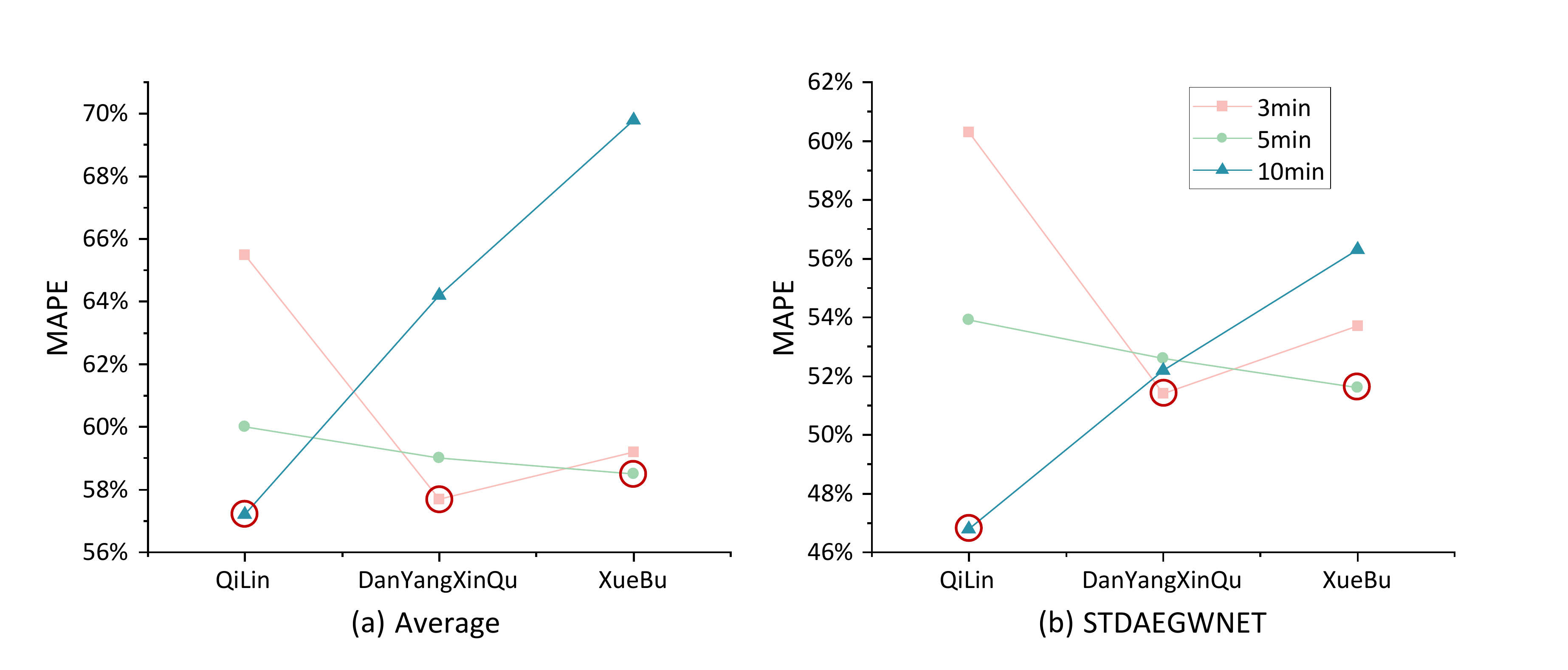}
  \caption{Comparison of MAPE across different sampling intervals. (a) Average MAPE of 14 models. (b) MAPE of the STDAEGWNET model.}
  \label{fig:mape_cmparison}
\end{figure}

The impact of sampling interval on model prediction performance is investigated by comparing the models on three interchanges—QiLin, DanYangXinQu, and XueBu—using 3 min, 5 min, and 10 min sampling intervals, with performance evaluated using MAPE, as shown in \figurename~\ref{fig:mape_cmparison}. The average performance of the models in Tables~\ref{tab:qilin_overall}, \ref{tab:danyangxinqu_overall}, and \ref{tab:xuebu_overall} shows that for the QiLin dataset, MAPE decreases as the sampling interval increases, from 65.5\% at 3 min to 57.2\% at 10 min, indicating that the 10 min interval is optimal. In contrast, for the DanYangXinQu dataset, MAPE increases from 57.7\% at 3 min to 64.2\% at 10 min, making the 3 min interval the best. The XueBu dataset performs best at a 5 min interval, with MAPE of 58.5\%, lower than 59.2\% at 3 min and 69.8\% at 10 min. Taking the STDMAEGWNET model as an example, its optimal MAPE values on the QiLin, DanYangXinQu, and XueBu datasets are 46.8\% at 10 min, 51.4\% at 3 min, and 51.6\% at 5 min, respectively, further confirming this trend. This variation directly reflects the unique physical and functional characteristics of each interchange as described in Table~\ref{details_of_selected_interchanges}. As a major ``urban-core'' hub connecting Nanjing's ring road and intercity expressways, QiLin manages massive but highly periodic commuting volumes. This strong macroscopic trend makes longer intervals (10 min) highly beneficial for smoothing out high-frequency micro-level noise (e.g., localized lane-changing). Conversely, DanYangXinQu acts as a critical transit junction between the major ``east-west'' Shanghai-Nanjing corridor and north-south routes. The traffic here involves more irregular, bursty freight and long-distance transit flows, resulting in frequent fluctuations where shorter intervals (3 min) are necessary to capture and retain critical dynamic information before it is over-smoothed. Finally, XueBu, functioning as a four-loop regional connector, represents an intermediate case where the standard 5 min interval optimally balances noise smoothing with the retention of regional traffic wave information.

\subsection{Ablation study}
\label{Ablation study}
\subsubsection{Pre-training strategy}
\label{Ablation on Pre-training Strategy}
The effectiveness and necessity of the proposed spatiotemporal encoding strategy are evaluated through three carefully designed model variants, which separately examine the roles of temporal and spatial dependencies. These variants help clarify how each component contributes to the overall reconstruction capability and predictive performance of STDAE.

\begin{itemize}
\item \textbf{TAE:} Removes the spatial conditioning input and reconstructs sequences using only TAE, thereby gauging the contribution of spatial context.
\item \textbf{SAE:} Removes the temporal conditioning input and reconstructs sequences using only SAE, thereby gauging the contribution of temporal context.
\item \textbf{w/o STDAE:} Removes the temporal and spatial conditioning inputs.
\end{itemize}

\begin{figure}[t]
  \centering
  \includegraphics[width=\textwidth]{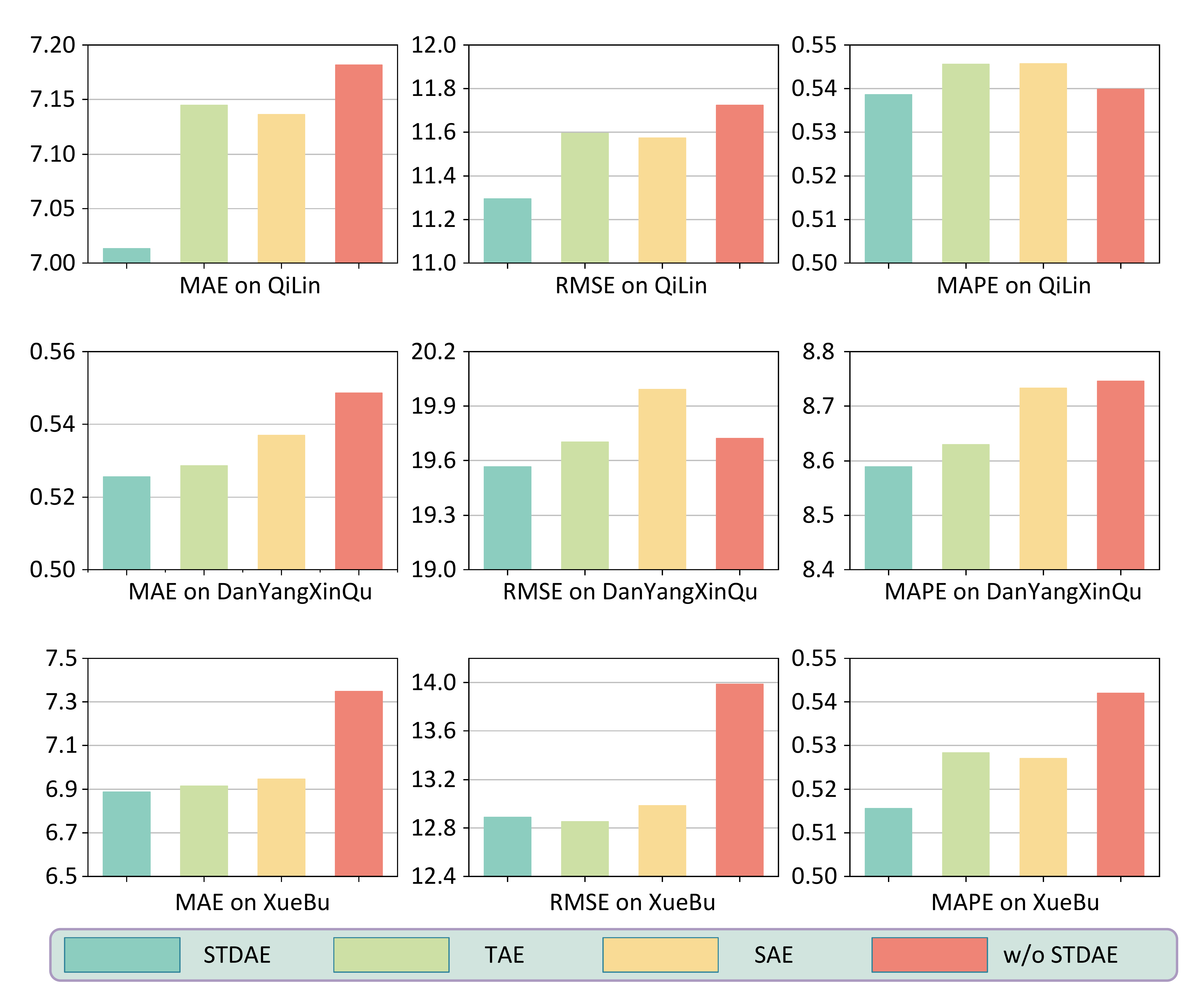}
  \caption{Comparison of STDAE and its ablated versions on three datasets with a 5 min sampling interval.}
  \label{fig:pretrain_ablation}
\end{figure}

The experimental results for the three datasets, using a 5 min sampling interval, are presented in \figurename~\ref{fig:pretrain_ablation}. STDAE consistently outperforms all ablated versions. This advantage mainly stems from its design of reconstructing ramp traffic flow by leveraging mainline traffic features. Compared with approaches that directly rely on historical ramp flow, this mechanism extracts richer contextual information from mainline data, thereby compensating for the absence of ramp information under “real-time blind spot” conditions. In the spatial dimension, mainline flow reflects the traffic environment upstream and downstream of the ramp. In the temporal dimension, mainline features capture the underlying causes of ramp flow fluctuations, such as upstream arrival waves and downstream capacity constraints. 

Furthermore, an in-depth analysis of the ablated variants reveals insightful patterns regarding the nature of traffic propagation. As observed, the temporal component (TAE) generally exhibits superior performance compared to the spatial component (SAE). This is primarily because highway traffic flow is heavily dominated by strong diurnal periodicity (e.g., morning and evening commuting peaks). TAE excels at capturing these robust temporal priors, providing a reliable baseline for overall volume trends. Conversely, spatial correlations between the mainline and specific ramps can sometimes be noisy or irregular due to complex lane-changing behaviors and diverse driver routing choices. A qualitative analysis of their respective failure modes further highlights their complementary roles: the SAE-only model often struggles to anticipate the onset of sudden peak flows, as it lacks the historical periodic context to predict temporal shifts before they fully manifest spatially. On the other hand, while the TAE-only model successfully anticipates the timing of peaks, it sometimes fails at accurate spatial distribution—for instance, misjudging exactly what proportion of an arriving mainline traffic wave will diverge into a specific off-ramp during a localized congestion event. By integrating both, STDAE successfully grounds the temporal periodicity with precise spatial distribution mapping, thereby significantly enhancing overall prediction accuracy and robustness.

\subsubsection{Downstream predictors}
\label{Ablation on downstream predictors}
The generality of the proposed STDAE framework is further examined by integrating it with five downstream predictors featuring diverse architectural backbones, including RNN, GCN+RNN, GCN+TCN, and Transformer:

\begin{itemize}
\item \textbf{STDAE-LSTM:} Using LSTM as the predictor.
\item \textbf{STDAE-STGCN:} Using STGCN as the predictor.
\item \textbf{STDAE-D2STGNN:} Using D2STGNN as the predictor.
\item \textbf{STDAE-STAEformer:} Using STAEformer as the predictor.
\item \textbf{STDAEGWNET:} Using GWNet as the predictor.
\end{itemize}

The experiments are conducted on three datasets with a 5 min sampling interval. Table~\ref{tab:ablation_predictor} presents the ablation results of STDAE across different downstream predictors. The experiments show that for all five downstream predictors, models enhanced with STDAE exhibit consistent and significant performance improvements across the three datasets.

\begin{table}[htbp]
\centering
\caption{Ablation Study of STDAE Across Different Downstream Predictors}
\label{tab:ablation_predictor}
\begin{threeparttable}
\begin{tabular}{l ccc}
\begin{tabularx}{\textwidth}{l *{3}{>{\centering\arraybackslash}X}}
\toprule
\multirow{2}{*}{Model} & QiLin & DanYangXinQu & XueBu \\
\cline{2-4}
& MAE/MAPE/RMSE & MAE/MAPE/RMSE & MAE/MAPE/RMSE \\
\midrule
LSTM & $7.68/12.63/0.56$ & $8.95/19.97/0.52$ & $7.43/13.89/0.57$ \\
STDAE-LSTM & $\mathbf{7.50}^{*}/\mathbf{12.34}^{*}/\mathbf{0.55}$ & $9.20/20.82/\mathbf{0.51}^{*}$ & $\mathbf{7.22}^{*}/\mathbf{13.43}^{*}/\mathbf{0.54}^{*}$ \\
\hline
STGCN & $7.45/12.09/0.55$ & $9.04/20.04/0.51$ & $7.23/13.25/0.55$ \\
STDAE-STGCN & $\mathbf{7.37}^{*}/\mathbf{11.85}^{*}/0.56$ & $\mathbf{8.94}^{*}/\mathbf{19.92}/0.53$ & $\mathbf{7.09}^{*}/\mathbf{13.11}^{*}/\mathbf{0.54}$ \\
\hline
D2STGNN & $7.34/12.02/0.54$ & $8.58/19.32/0.50$ & $6.98/12.95/0.50$ \\
STDAE-D2STGNN & $\mathbf{7.21}^{*}/\mathbf{11.75}^{*}/\mathbf{0.53}$ & $\mathbf{8.51}/\mathbf{19.26}/0.51$ & $\mathbf{6.89}^{*}/\mathbf{12.77}/0.51$ \\
\hline
STAEformer & $7.39/11.97/0.55$ & $9.05/20.31/0.53$ & $7.12/13.01/0.53$ \\
STDAE-STAEformer & $\mathbf{7.26}^{*}/\mathbf{11.73}^{*}/0.55$ & $\mathbf{8.80}^{*}/\mathbf{19.64}^{*}/0.54$ & $\mathbf{7.11}/\mathbf{12.97}/0.54$ \\
\hline
GWNET & $7.18/11.72/0.54$ & $8.75/19.72/0.55$ & $7.35/13.99/0.54$ \\
STDAEGWNET & $\mathbf{7.01}^{*}/\mathbf{11.29}^{*}/\mathbf{0.54}$ & $\mathbf{8.59}^{*}/\mathbf{19.57}/\mathbf{0.53}$ & $\mathbf{6.89}^{*}/\mathbf{12.89}^{*}/\mathbf{0.52}^{*}$ \\
\bottomrule
\end{tabularx}
\end{tabular}
\begin{minipage}{\linewidth}
\footnotesize
\textbf{Note:} Bold values indicate better performance (lower values) compared to baseline models. $^*$ indicates statistically significant improvement ($p < 0.05$).
\end{minipage}
\end{threeparttable}
\end{table}

Specifically, STDAE-LSTM shows reductions in MAE, MAPE, and RMSE compared to the baseline LSTM on the QiLin and XueBu datasets, indicating that the representations generated by STDAE effectively enhance the predictive capability of traditional sequential models. Similarly, when combined with graph neural network predictors, STDAE achieves superior performance across all datasets compared to their baseline versions, demonstrating that STDAE can extract rich spatiotemporal features from mainline traffic flows and convey them to downstream models. For more complex Transformer architectures, STDAE also brings noticeable performance gains. Overall, STDAE consistently and significantly improves performance across various predictors, validating that the mainline feature representations it generates are highly generalizable and robust.

These results indicate that STDAE not only independently enhances model performance but also serves as a general enhancement module, compatible with downstream predictors of different architectures, thereby improving the overall effectiveness of traffic flow forecasting.

\subsubsection{Ramp Graph Structure}
\label{Ablation on Graph Structure}

To validate the effectiveness of the ramp graph design, we conduct an ablation study comparing different prior graph constructions. The key challenge in spatial modeling lies in providing a meaningful structural prior while avoiding the introduction of noisy correlations from spatially or functionally unrelated ramps. Since the downstream predictor GWNet incorporates a data-driven adaptive graph learning mechanism, it is necessary to examine how different prior topologies interact with this adaptive component.

We evaluate the following three configurations under the 5\,min sampling interval:

\begin{itemize}
    \item \textbf{Distance-based + Adaptive Graph (Ours):} The prior adjacency matrix connects physically proximate ramps, combined with the adaptive graph mechanism to capture latent data-driven dependencies.
    \item \textbf{Fully Connected + Adaptive Graph:} A fully connected prior graph is assumed, together with the adaptive graph mechanism.
    \item \textbf{Distance-based w/o Adaptive Graph:} The prior adjacency matrix connects physically proximate ramps, while the adaptive graph mechanism is disabled, relying solely on the predefined physical topology.
\end{itemize}

\begin{table}[htbp]
\centering
\caption{MAE performance under different ramp graph structures (5 min sampling interval).}
\label{tab:graph_ablation}
\begin{tabular}{l c c c c}
\toprule
\textbf{Prior Graph Structure} & \textbf{Adaptive Graph} & \textbf{QiLin} & \textbf{DanYangXinQu} & \textbf{XueBu} \\
\midrule
Fully Connected & $\checkmark$ & 7.15 & 8.64 & 6.92 \\
Distance-based & $\times$ & 7.16 & 8.73 & 7.04 \\
\textbf{Distance-based (Ours)} & $\checkmark$ & \textbf{7.01} & \textbf{8.59} & \textbf{6.89} \\
\bottomrule
\end{tabular}
\end{table}

As shown in Table~\ref{tab:graph_ablation}, the combination of a distance-based prior with an adaptive graph achieves the best performance across all datasets. The ``Fully Connected + Adaptive Graph'' configuration underperforms compared to the distance-based prior, indicating that enforcing a fully connected topology introduces unnecessary and potentially noisy correlations among distant or weakly related ramps. This increases the burden on the adaptive mechanism, which must allocate additional capacity to suppress irrelevant connections.

In contrast, the ``Distance-based w/o Adaptive Graph'' variant yields the largest prediction errors, demonstrating that physical proximity alone is insufficient to capture complex and non-local dependencies inherent in traffic dynamics. These results confirm that a reasonable structural prior, when complemented by data-driven adaptive learning, strikes an effective balance between inductive bias and flexibility. By initializing the model with a physically meaningful topology and allowing dynamic refinement, the proposed configuration effectively filters out irrelevant noise while capturing latent correlations, leading to optimal predictive performance.

\subsubsection{Hyperparameter study}
\label{Ablation on Hyperparameter}
\begin{figure}[t]
    \centering
    \includegraphics[width=\linewidth]{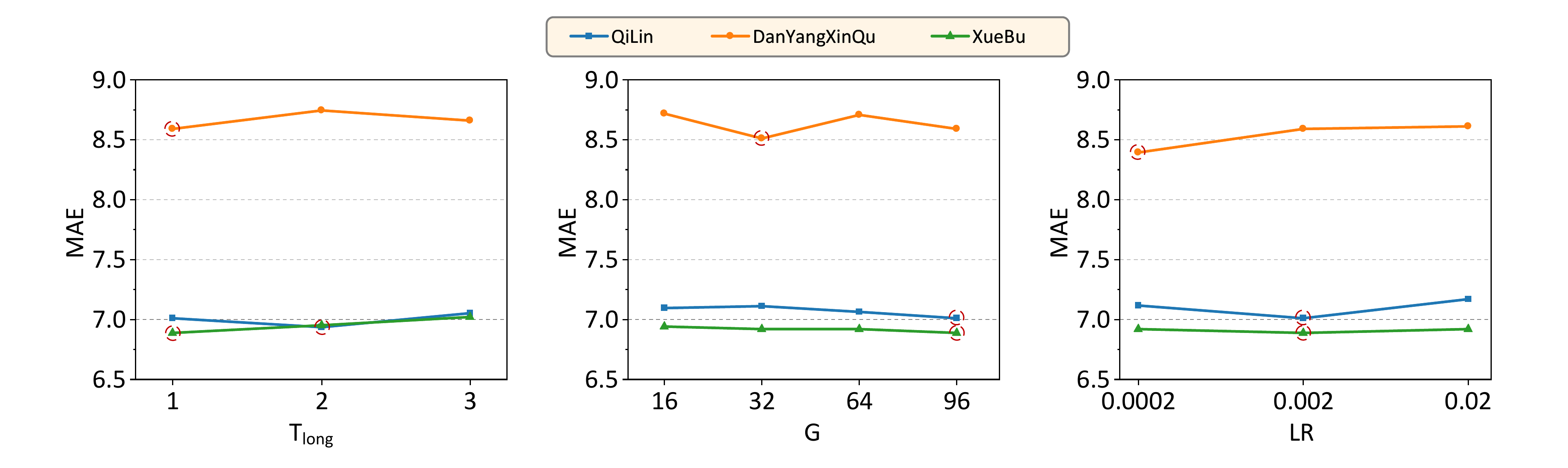}
    \caption{MAE performance under manual tuning of hyperparameters ($T_{\text{long}}$, $G$ and learning rate) across three datasets with a 5 min sampling interval.}
    \label{fig:HyperparameterSearch}
\end{figure}

To determine the optimal configuration for our model, we conduct a manual hyperparameter search focusing on three critical parameters: $T_{\text{long}}$, LR, and $G$. The evaluation is performed on the QiLin, DanYangXinQu, and XueBu datasets using a 5 min sampling interval. The performance variations in terms of MAE are illustrated in \figurename~\ref{fig:HyperparameterSearch}. The final hyperparameters are selected based on the configuration that yields the minimum MAE on the majority (two out of three) of the datasets.

\textbf{Impact of $T_{\text{long}}$:} We evaluate $T_{\text{long}} \in \{1, 2, 3\}$ days to determine the optimal historical sequence length for the pre-training stage. As shown in the experimental results, while $T_{\text{long}} = 2$ achieves a slightly better MAE on the QiLin dataset (6.94 compared to 7.01), setting $T_{\text{long}} = 1day$ results in the lowest MAE on both the DanYangXinQu (8.59) and XueBu (6.89) datasets. Furthermore, the total MAE across all three datasets is the lowest at $T_{\text{long}} = 1day$ (22.49). This suggests that while extending the historical window provides more data, excessively long sequences may introduce irrelevant temporal noise and outdated traffic fluctuations. Such noise can interfere with the model's ability to focus on the most immediate and critical spatiotemporal dependencies. Therefore, $T_{\text{long}} = 1day$ is selected to optimally capture fundamental diurnal traffic patterns while balancing computational efficiency and prediction accuracy.

\textbf{Impact of LR:} We test initial learning rates of 0.02, 0.002, and 0.0002. A learning rate of 0.002 achieves the best performance on the QiLin and XueBu datasets. Although a smaller learning rate of 0.0002 performs better on the DanYangXinQu dataset, 0.002 provides consistent convergence across the majority of the scenarios and is thus selected.

\textbf{Impact of $G$:} The embedding dimension controls the representational capacity of the model. We explore $G \in \{16, 32, 64, 96\}$. The results indicate that a larger dimension of $G = 96$ yields the lowest MAE on the QiLin and XueBu datasets. While $G = 32$ shows a lower MAE on the DanYangXinQu dataset, $G = 96$ is adopted as the final setting due to its superior performance on the other two datasets, allowing the model to capture more complex spatiotemporal dependencies.

\subsection{Robustness on Missing Data}
\label{Performance under masking}

\begin{figure}[t]
  \centering
  \includegraphics[width=\textwidth]{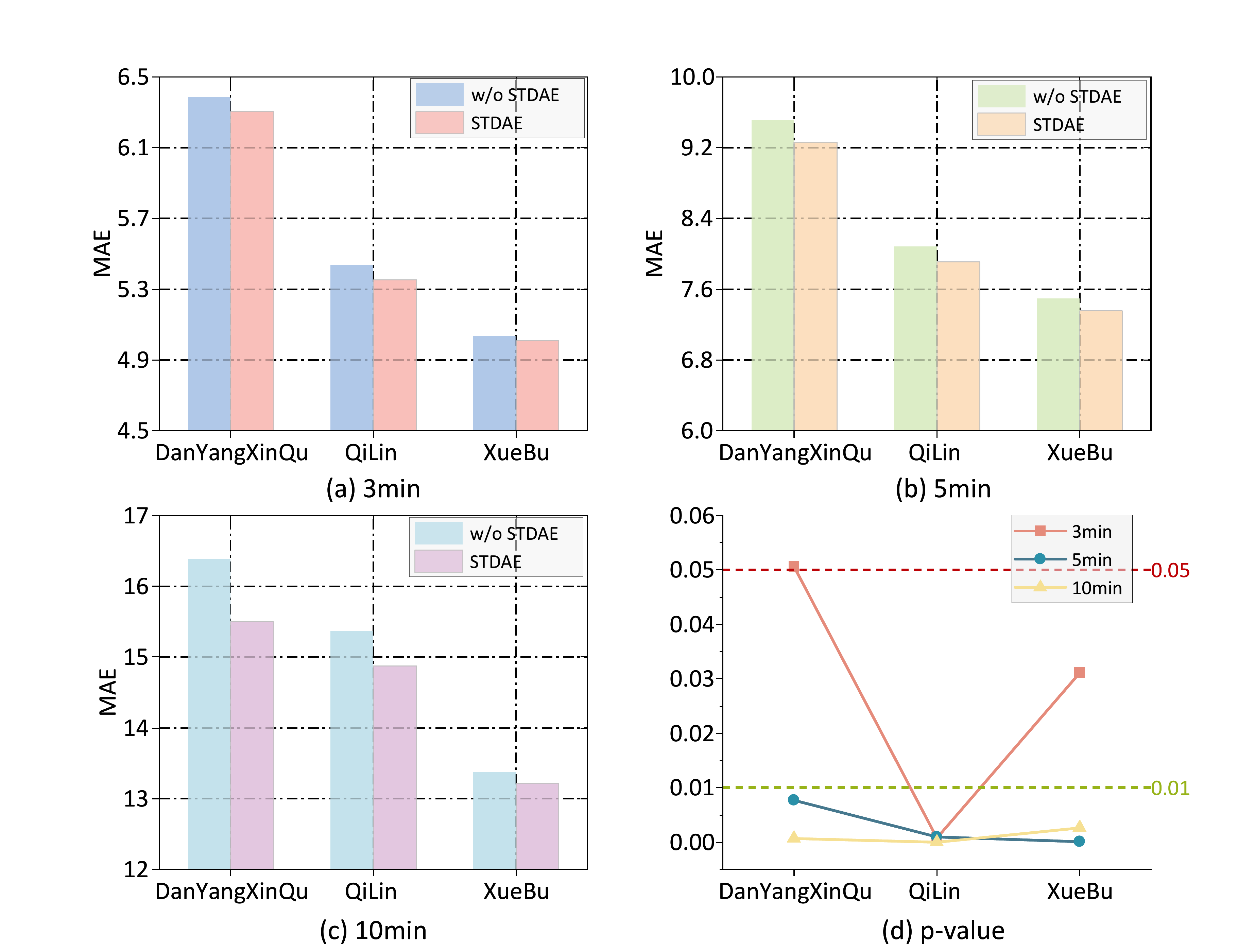}
  \caption{MAE Comparison of STDAEGWNET with and without STDAE under Different Sampling Intervals and Significance Analysis. (a) 3 min. (b) 5 min. (c) 10 min. (d) Significance analysis.}
  \label{fig:mask_ablation}
\end{figure}

To evaluate the robustness of STDAEGWNET under missing mainline data conditions, structured directional and temporal masking strategies are consistently applied to the mainline data during the pre-training, training, and testing phases to simulate severe real-world data loss scenarios. This unified masking strategy serves both as a learning mechanism and as an evaluation protocol, enabling the model to adapt to contiguous sensor or communication failures while ensuring a fair robustness assessment. Specifically, the directional mask removes data from two eastbound mainline ETC gantries, simulating a regional sensor outage. The temporal mask removes the last 6 steps out of every 12 time steps, simulating a continuous communication interruption. Experiments are conducted on three datasets with sampling intervals of 3 min, 5 min, and 10 min. Models with and without STDAE are compared, and the MAE results are presented in \figurename~\ref{fig:mask_ablation}. The results demonstrate that, across all datasets and sampling intervals, incorporating STDAE significantly enhances prediction accuracy under simulated missing data conditions. Compared with baseline models without STDAE, the enhanced models achieve a consistent performance gain, with an average MAE reduction of approximately 2.23\%. Moreover, most improvements are statistically significant ($p < 0.01$).

\begin{figure}[t]
  \centering
  \includegraphics[width=\textwidth, keepaspectratio]{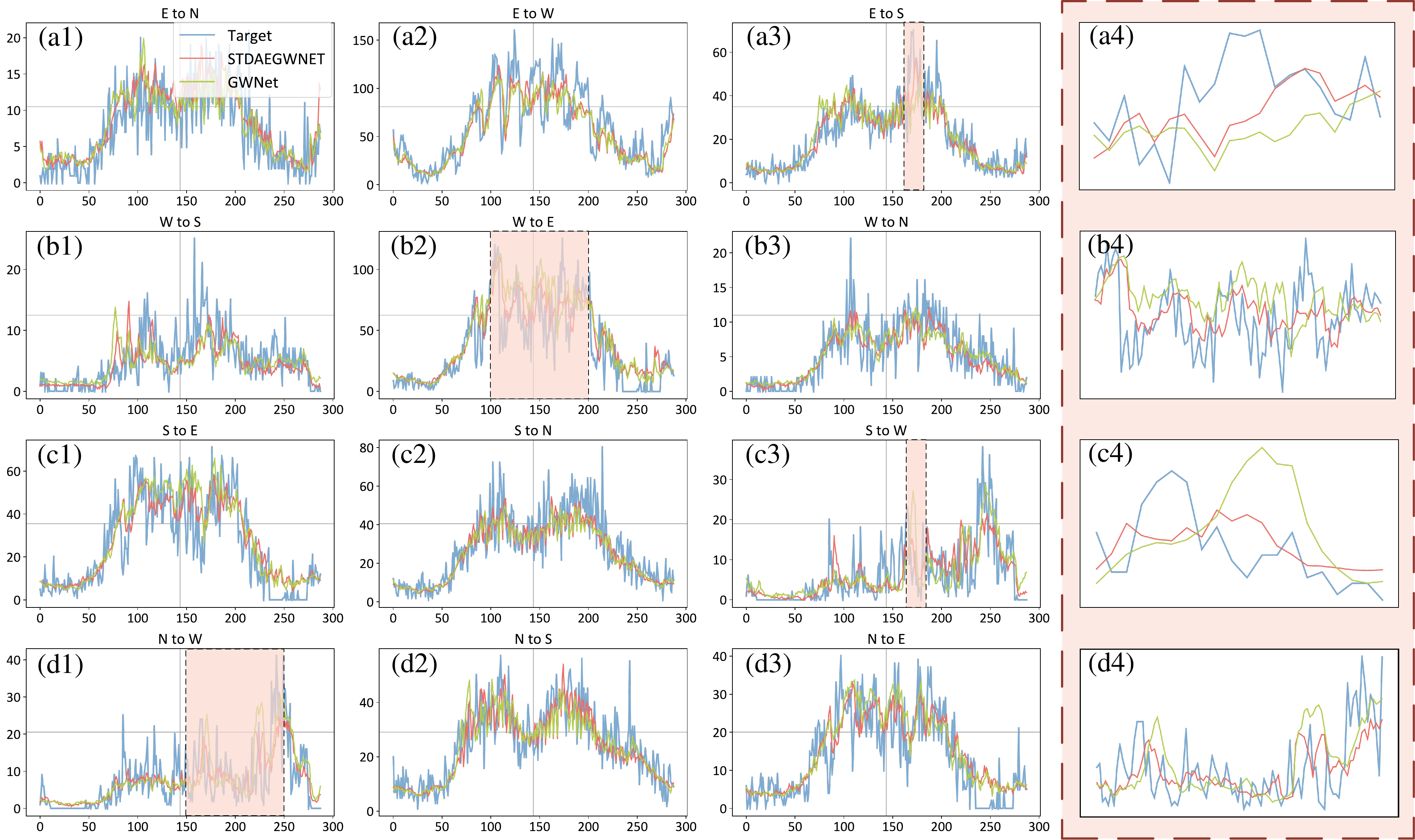}
  \caption{Ramp-level Prediction of the QiLin Interchange with a 5 min Sampling Interval (step = 3). Subfigure labels correspond to the source direction: (a) East, (b) West, (c) South, and (d) North. Turning movements are indicated as follows: (1) right-turn, (2) straight, and (3) left-turn. Various traffic dynamics: (a4) an upward trend, (b4) a fluctuating trend, (c4) a downward trend, and (d4) a fluctuating upward trend.}
  \label{fig:prediction_visualization}
\end{figure}

\subsection{Efficiency study}
\label{Efficiency Study}
To comprehensively evaluate the practical applicability and computational overhead of the proposed framework, we conduct a detailed runtime and complexity analysis. We compare the training time per epoch, test time, number of learnable parameters, and peak GPU memory usage of our models against several representative baselines. The efficiency test is performed on the QiLin dataset with a 5 min sampling interval, and all models are executed on a workstation equipped with a single NVIDIA RTX 4090D (24GB) GPU. The quantitative results are summarized in Table~\ref{tab:efficiency}.

\begin{table}[t]
\centering
\caption{Computational complexity and runtime analysis on the QiLin dataset (5 min interval).}
\label{tab:efficiency}
\begin{tabular}{l c c c c}
\toprule
\textbf{Model} & \textbf{Train Time (s/epoch)} & \textbf{Test Time (s)} & \textbf{Parameters} & \textbf{Peak GPU Memory (MB)} \\
\midrule
STDAE-SAE (Pre-train) & 4.32 & 2.64 & 574,764 & 724.73 \\
STDAE-TAE (Pre-train) & 4.39 & 2.61 & 574,764 & 704.30 \\
\midrule
GWNET (Baseline)      & 4.39 & 2.67 & 127,068 & 94.01  \\
STAEformer (Baseline) & 1.27 & 0.07 & 349,084 & 214.61 \\
D2STGNN (Baseline)    & 4.92 & 0.26 & 86,698  & 196.76 \\
\midrule
\textbf{STDAEGWNET (Ours)} & 5.16 & 2.88 & 300,316 & 135.82 \\
\bottomrule
\end{tabular}
\end{table}

As shown in Table~\ref{tab:efficiency}, the pre-training modules (STDAE-SAE and STDAE-TAE) require approximately 574K parameters and consume peak GPU memories of 724.73 MB and 704.30 MB, respectively. Their training times are relatively fast, taking roughly 4.3 to 4.4 seconds per epoch. Although the proxy reconstruction mechanism introduces additional computational parameters during the pre-training phase, it is crucial to emphasize that this is a one-time, offline process.

In the downstream forecasting stage, the integrated STDAEGWNET model demonstrates high operational efficiency. It contains approximately 300K parameters and requires only 135.82 MB of peak GPU memory. Compared to the original GWNET backbone, STDAEGWNET only modestly increases the training time (from 4.39s to 5.16s) and test time (from 2.67s to 2.88s), while keeping the GPU memory footprint significantly lower than other baselines like STAEformer (214.61 MB) and D2STGNN (196.76 MB). This minimal computational burden indicates that the learned spatiotemporal representations injected by STDAE efficiently enhance prediction accuracy without compromising real-time inference capability.

\subsection{Accuracy Across Ramps}
\label{Accuracy Across Ramps}

To further evaluate model performance, a micro-level analysis of prediction accuracy for individual ramps is conducted. This section examines the improvement of STDAEGWNET over its backbone network, GWNet, in capturing the dynamics of specific ramp traffic flows. The results from the QiLin interchange with a 5 min sampling interval are selected, and the predictions for the first day in the test set are visualized.

As shown in \figurename~\ref{fig:prediction_visualization}, the figure intuitively presents the comparison between the predicted curves and the ground truth for STDAEGWNET and GWNet at step = 3. Visually, the prediction curves of STDAEGWNET closely follow the fluctuations of the actual traffic, particularly at the peaks and troughs of traffic flow. Four time segments from \figurename~\ref{fig:prediction_visualization} (a3), (b2), (c3), and (d1) are selected to illustrate the prediction results, as shown in \figurename~\ref{fig:prediction_visualization} (a4)-(d4). These correspond to traffic flow trends of increase, fluctuation, decrease, and uctuating increase, respectively. The blue curve represents the background ramp traffic flow, the green curve represents the GWnet prediction, and the red curve represents the prediction with STDAE added. Intuitively, adding STDAE more accurately predicts traffic flow trends of increase and decrease. It also better matches the actual traffic flow compared to GWnet in the case of traffic fluctuations.

For quantitative comparison, the performance metrics for each turning ramp over the entire test set are calculated, and the detailed results are presented in Table~\ref{tab:ramp_comparison}. The quantitative results clearly confirm the conclusions from the visualization analysis. From the table, it can be seen that out of 12 ramps, STDAEGWNET outperforms the original GWNet in 10 ramps in terms of overall prediction performance. Particularly for major directions with higher traffic volumes and more challenging predictions, such as ``W to E'' and ``E to W'', STDAEGWNET shows a significant advantage, with MAE reduced by approximately 9.0\% and 4.9\%, respectively, and RMSE also decreased accordingly. This indicates that the model predicts complex traffic patterns more accurately and robustly. 

\begin{table}[t]
\centering
\caption{Ramp-level Performance Comparison between STDAEGWNET and GWNet of the QiLin Interchange with a 5 min Sampling Interval}
\label{tab:ramp_comparison}
\renewcommand{\arraystretch}{1.2} 
\begin{tabular}{
>{\centering\arraybackslash}m{2.2cm} 
>{\centering\arraybackslash}m{1.5cm} 
>{\centering\arraybackslash}m{1.5cm} 
>{\centering\arraybackslash}m{1.5cm} 
>{\centering\arraybackslash}m{1.5cm} 
>{\centering\arraybackslash}m{1.5cm} 
>{\centering\arraybackslash}m{1.5cm} 
>{\centering\arraybackslash}m{2.2cm} 
}
\toprule
\multirow{2}{*}{\textbf{Direction}} & \multicolumn{3}{c}{\textbf{STDAEGWNET}} & \multicolumn{3}{c}{\textbf{GWNet}} & \multirow{2}{*}{\textbf{Better Model}} \\
\cmidrule(lr){2-7}
 & MAE & MAPE(\%) & RMSE & MAE & MAPE(\%) & RMSE & \\
\midrule
E to N & 2.90 & 53.29 & 3.84 & 2.84 & 50.42 & 3.78 & GWNet \\
E to W & 15.61 & 42.41 & 21.14 & 16.42 & 42.94 & 22.22 & STDAEGWNET \\
E to S & 5.64 & 35.75 & 8.12 & 5.82 & 34.84 & 8.38 & STDAEGWNET \\
W to S & 2.84 & 55.13 & 4.32 & 2.94 & 55.73 & 4.38 & STDAEGWNET \\
W to E & 14.11 & 66.34 & 19.55 & 15.50 & 70.78 & 21.99 & STDAEGWNET \\
W to N & 2.20 & 43.64 & 3.02 & 2.24 & 43.91 & 3.08 & STDAEGWNET \\
S to E & 9.79 & 63.65 & 13.08 & 9.87 & 63.00 & 13.54 & STDAEGWNET \\
S to N & 7.01 & 30.60 & 9.72 & 7.11 & 31.85 & 9.84 & STDAEGWNET \\
S to W & 4.34 & 88.80 & 6.30 & 4.58 & 86.21 & 6.69 & STDAEGWNET \\
N to W & 4.30 & 85.99 & 6.24 & 4.47 & 89.90 & 6.51 & STDAEGWNET \\
N to S & 6.13 & 28.84 & 8.50 & 6.39 & 29.05 & 8.98 & STDAEGWNET \\
N to E & 5.85 & 60.52 & 7.61 & 5.74 & 60.29 & 7.59 & GWNet \\
\bottomrule
\end{tabular}%
\end{table}

As shown in the first and last rows of Table \ref{tab:ramp_comparison}, GWNet exhibits a slight performance advantage for the “E to N” and “N to E” ramps, with MAE of 2.84 and 5.74, MAPE of 50.42\% and 60.29\%, and RMSE of 3.78 and 7.59, respectively. Analysis of the traffic characteristics reveals that these ramps likely have relatively low traffic volumes or simpler patterns, where the deep spatiotemporal context injected by STDAE pre-training does not play a critical role, and a strong direct prediction model can already achieve good performance. Overall, the fine grained comparative analysis conducted at the ramp level clearly demonstrates the practical value of the STDAE pre-training framework. By learning deep mapping relationships between the mainline and ramps, the framework significantly enhances the performance of the backbone network at the micro level, particularly in complex traffic scenarios, leading to substantial improvements in predictive accuracy.

\section{Conclusion}
\label{Conclusion}
This study proposes the STDAE framework, a two-stage approach for ramp traffic flow prediction at highway interchanges. It first performs proxy reconstruction pre-training and then integrates the learned representations with downstream predictors. In the pre-training stage, STDAE reconstructs ramp flows from mainline traffic sequences collected by ETC systems, enabling its spatial and temporal branches to capture heterogeneous but complementary features. After feature fusion, the framework provides downstream predictors with rich long-range contextual information.

Comprehensive experiments on multi-week ETC datasets at three real-world cloverleaf interchanges and under multiple sampling intervals confirm the effectiveness of the proposed method. GWNet is adopted as the downstream predictor, and the combined model STDAEGWNET achieves the best overall performance across MAE, MAPE, and RMSE metrics among thirteen competitive baselines on all three datasets, with average ranks of 1.00, 2.11, and 1.44 on the QiLin, DanYangXinQu, and XueBu datasets, respectively. Moreover, at a 3 min sampling interval, the model achieves the lowest mean absolute errors of 4.89, 5.61, and 4.58 on the QiLin, DanYangXinQu, and XueBu datasets, respectively, demonstrating a significant accuracy advantage over other methods. Compared with advanced models that rely on historical ramp flow sequences, STDAEGWNET demonstrates comparable predictive capability while relying solely on mainline data. Since STDAE relies solely on widely available mainline ETC data, it enables fine-grained interchange control even in the absence of ramp detectors, without requiring hardware retrofitting. Extensive ablation studies further reveal that the proposed STDAE generates informative representations that effectively capture spatial and temporal dependencies between mainline and ramp flows. The learned embeddings are architecture-agnostic and can be seamlessly injected into other sequence or graph models with minimal modifications, serving as a plug-and-play enhancement for diverse forecasting pipelines. In addition, STDAEGWNET remains robust and achieves favorable performance even under scenarios with partially missing mainline data, maintaining stable competitiveness and strong generalizability across all three hub datasets.

In summary, the experimental findings validate the effectiveness of the STDAEGWNET framework.  It provides a practical solution to the challenge of unavailable ramp data under real-time blind spot conditions. However, several limitations in the current study must be explicitly acknowledged. First, the proxy reconstruction mechanism has inherent limitations; its effectiveness may be compromised when ramp flows are highly irregular or exhibit exceptionally weak correlations with mainline traffic. Second, due to the relatively short temporal span of our dataset (a 17-day training period), there is a potential risk of temporal overfitting, particularly concerning the model's ability to capture long-term seasonal variations. Third, the current method heavily depends on the assumption of an accurate and reliable mainline-ramp topological structure. Finally, there is a lack of nationwide transferability tests.

Future work will address these limitations by incorporating multi-source data (e.g., weather and incident reports) to handle irregular flows, expanding the training datasets to span multiple seasons, and relaxing the dependence on exact topologies through more advanced dynamic graph learning. Most importantly, validating the framework's generalizability and transferability across diverse interchange environments at a nationwide network level remains a critical direction for our future research.




\bibliographystyle{elsarticle-num}
\bibliography{references_r1}

\end{document}